\documentclass[10pt, journal, web]{article}
\usepackage{amsmath, amssymb, amsfonts, amsthm} 
\usepackage{graphicx}  
\usepackage{color, xcolor}  
\usepackage{multirow, booktabs, array}  
\usepackage{adjustbox} 
\usepackage{mathtools}  
\usepackage{epsfig}  
\usepackage{setspace}  
\usepackage[english]{babel} 
\usepackage{lineno}  
\usepackage{booktabs}  
\usepackage{multirow}  
\usepackage{array}  
\usepackage{subcaption}  
\usepackage{lscape}
\usepackage{slashbox}
\usepackage{multirow}
\usepackage[normalem]{ulem}
\useunder{\uline}{\ul}{}

\usepackage{bbm}
\usepackage[ruled,vlined]{algorithm2e}
\usepackage{pifont}
\usepackage{xcolor}
\usepackage{hyperref}
\usepackage{booktabs}
\hypersetup{
   colorlinks=true,
    linkcolor=blue,
   citecolor=blue,
    urlcolor=blue
}
\usepackage{makecell}
\usepackage{tabularx}
\usepackage{enumitem}

\newcommand{\cvect}{\mbox{\bf c}}

\newcommand{\svect}{\mbox{\bf s}}

\newcommand{\yvect}{\mbox{\bf y}}
\newcommand{\zvect}{\mbox{\bf z}}

\makeatletter
\usepackage{authblk}  
\usepackage{tabularx}

\makeatletter
\renewcommand\AB@affilsepx{, \protect\Affilfont}
\makeatother

\usepackage{tabularx}
\providecommand{\keywords}[1]{%
  \small	
  \textbf{\textit{Keywords---}} #1%
}

\begin{document}

\title{\textbf{OSCS-SupCon: Orthogonal Sigmoid-based Common and Style Supervised Contrastive Learning for Robust Feature Disentanglement}}
\author[1]{B. Wang}
\author[1, 2]{F. Dornaika\thanks{Corresponding author}}
\affil[1]{\textit{University of the Basque Country}}
\affil[2]{\textit{IKERBASQUE}}

\affil[ ]{

\small\texttt{bwang001@ikasle.ehu.eus, fadi.dornaika@ehu.eus}}
\date{}
\maketitle
\begin{abstract}
Supervised Contrastive Learning (SupCon) has achieved remarkable success by explicitly modeling pairwise relationships among samples. However, existing SupCon-based methods face two critical limitations: negative-sample dilution arising from the conventional InfoNCE loss, and feature-space entanglement due to the absence of explicit constraints to separate category-relevant (common) from category-irrelevant (style) features. These issues severely undermine feature discriminability and generalization.

To address these challenges, we propose OSCS-SupCon (Orthogonal Sigmoid-based Common and Style Supervised Contrastive Learning), a novel framework that integrates a sigmoid-based pairwise contrastive loss with explicit orthogonality constraints. Specifically, we introduce a sigmoid-based contrastive loss function with two learnable parameters—temperature and bias—which adaptively adjusts pairwise decision boundaries to mitigate negative-sample dilution. In addition to the style-distance constraint, we enforce rigorous orthogonality between common and style feature subspaces via a  linear projection with ReLU nonlinearity. This design thoroughly eliminates overlapping and ambiguous style representations. Extensive experiments on six benchmark datasets show that OSCS-SupCon consistently outperforms state-of-the-art supervised contrastive methods across various backbone encoders. Notably, on the challenging fine-grained CUB200-2011 dataset with a ResNet-18 backbone, our method achieves a 3.4\% accuracy improvement over CS-SupCon~\cite{dornaika2025deep}, demonstrating superior robustness and generalization. Ablation studies further validate the contribution of each proposed component.
\end{abstract}

\keywords{Supervised contrastive learning, Sigmoid contrastive loss, Orthogonality constraint, Feature disentanglement, Style distance constraint, Fine-grained classification}

\section{Introduction}

In recent years, metric learning and contrastive learning have become essential techniques for directly extracting discriminative features from raw data, significantly advancing various tasks such as image classification, face recognition, and zero-shot learning~\cite{DARBAN2025,pmlr-v162-deng22c,Bao2023,pmlr-v139-roth21a}. Typically, these methods project samples into a low-dimensional embedding space, reducing distances among similar-class samples while increasing distances among dissimilar-class samples~\cite{zhu2021visual,Gonzalez-Zapata_2022_CVPR,WANG2024}. Particularly, with the rapid evolution of deep learning, shallow embedding-based approaches have gradually been superseded by powerful deep metric learning frameworks, demonstrating significantly improved feature representation and generalization performance~\cite{Wang2022}. Among these methods, Supervised Contrastive Learning (SupCon)~\cite{khosla2020supervised}, which explicitly models pairwise relationships among samples, has shown remarkable success, especially in fine-grained image recognition and large-scale classification tasks~\cite{Li2023,ZHANG2023}.

However, despite SupCon's remarkable success, inherent limitations in its contrastive loss function and embedding structure have become increasingly apparent. Specifically, classical SupCon methods generally utilize a unified embedding space induced by the InfoNCE loss~\cite{oord2018representation}, simultaneously comparing each anchor against multiple negative samples within a batch. This formulation tends to dilute the contributions of critical negative samples, diminishing discriminative feature learning and ultimately limiting model generalization, particularly in fine-grained classification scenarios with subtle inter-class differences. Moreover, these methods neglect explicit disentanglement of category-relevant (common) and category-irrelevant (style) features, leading to severe feature-space entanglement. Early contrastive methods, such as IE loss~\cite{WU2018}, N-pair loss~\cite{Sohn2016}, and Multi-similarity loss~\cite{Wang2019}, laid the foundational principles for improving intra-class compactness and inter-class separation, significantly influencing modern supervised contrastive frameworks. Additionally, self-supervised contrastive approaches like SimCLR~\cite{Chen2020} and BYOL~\cite{Grill2020} further advanced representation learning through effective use of augmented views or removing negative samples entirely. These pioneering methods highlight the critical importance of effectively leveraging negative samples and explicitly managing intra-class variations to enhance discriminative feature learning, motivating the design of our proposed framework. Although the recently proposed CS-SupCon~\cite{dornaika2025deep} explicitly partitions deep features into common and style subspaces and introduces an explicit style-distance constraint, it still relies on the traditional InfoNCE-based contrastive loss~\cite{oord2018representation}, failing to effectively address the negative-sample dilution issue and feature subspace coupling simultaneously.

\begin{figure}[htbp]
    \centering
\includegraphics[width=1.1\textwidth]{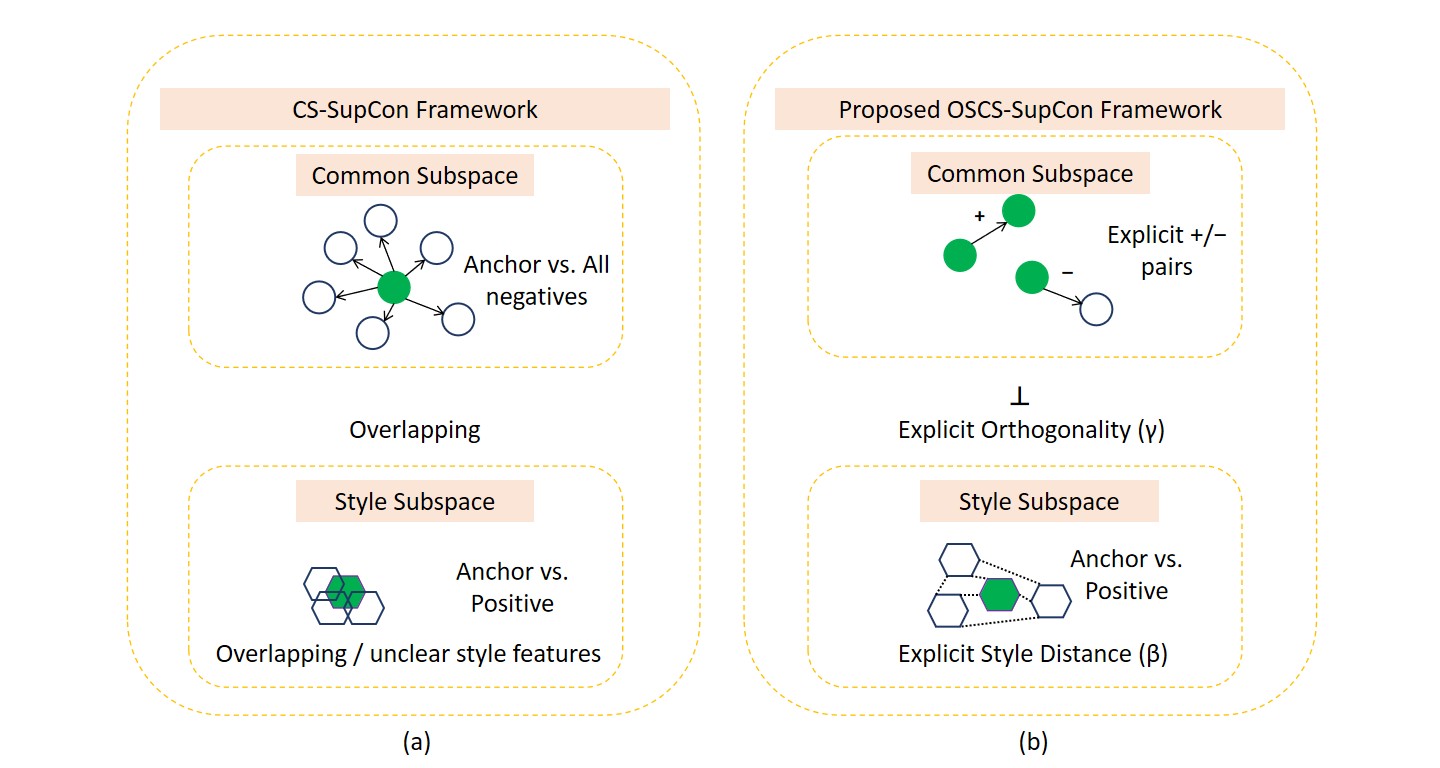}
    \caption{Motivation and key innovations of our proposed OSCS-SupCon method. (a) Existing supervised contrastive frameworks employ a unified InfoNCE-based loss, causing negative sample dilution, inadequate management of style variations, and significant feature-space coupling. (b) Proposed OSCS-SupCon explicitly introduces a sigmoid-based pairwise contrastive loss to mitigate negative sample dilution, an explicit style-distance constraint to handle intra-class style variations robustly, and explicit orthogonality constraints to rigorously decouple common and style feature subspaces.}
    \label{fig:OSCS_intro}
\end{figure}

As illustrated in Figure~\ref{fig:OSCS_intro}(a, top), this ``all-to-all'' comparison inevitably exacerbates the dilution of critical negative sample information. Additionally, as explicitly depicted in Figure~\ref{fig:OSCS_intro}(a, bottom), the absence of rigorous orthogonality constraints between common and style subspaces in CS-SupCon results in ambiguous and overlapping style representations and significant feature subspace coupling, thus reducing the interpretability and robustness of the learned representations. Similarly, SelfCon~\cite{baeSelfContrastiveLearningSingleviewed2022} simplifies data augmentation strategies but also adopts a unified InfoNCE-based embedding space, encountering similar issues.

Consequently, current supervised contrastive learning approaches urgently require addressing the following critical limitations clearly illustrated in Figure~\ref{fig:OSCS_intro}(a):

\textbf{Negative Sample Dilution}: Existing InfoNCE-based methods dilute the discriminative contribution of essential negative samples due to simultaneous anchor-negative comparisons (Figure~\ref{fig:OSCS_intro}(a, top)).

\textbf{Inadequate Handling of Style Variations}: Current methods lack explicit constraints to manage intra-class style variations, causing ambiguous and overlapping style representations (Figure~\ref{fig:OSCS_intro}(a, bottom)).

\textbf{Insufficient Feature Disentanglement}: Most existing methods implicitly handle feature disentanglement, lacking explicit orthogonality constraints to rigorously separate common and style features (also reflected by overlapping regions in Figure~\ref{fig:OSCS_intro}(a, bottom)).

To comprehensively address these challenges, we propose Orthogonal Sigmoid-based Common and Style Supervised Contrastive Learning (OSCS-SupCon), as illustrated in Figure~\ref{fig:OSCS_intro}(b). Specifically, OSCS-SupCon replaces the traditional InfoNCE-based loss with a novel sigmoid-based pairwise contrastive loss, effectively alleviating negative sample dilution by adaptively adjusting decision boundaries through two learnable parameters, temperature \( t \) and bias \( b \). Moreover, we incorporate the explicit style-distance penalty proposed by CS-SupCon~\cite{dornaika2025deep} to robustly manage intra-class style variations. Crucially, OSCS-SupCon enforces explicit orthogonality constraints between common and style subspaces through an additional linear projection with ReLU activation, ensuring rigorous feature disentanglement and significantly enhancing feature discriminability and robustness.

Extensive experiments and rigorous statistical validation (paired Student's t-test, \( p<0.05 \)) conducted across multiple public datasets, utilizing various backbone encoders including classic CNN encoders and advanced Transformers, clearly demonstrate that OSCS-SupCon consistently surpasses existing supervised contrastive methods. Notably, our method particularly excels in fine-grained classification tasks, validating its substantial generalization capability and robustness.

The main contributions of this paper are summarized as follows:

\begin{itemize}
    \item \textbf{Sigmoid-based Contrastive Loss Design:} We explicitly introduce a sigmoid-based pairwise contrastive loss into supervised contrastive learning, significantly alleviating negative sample dilution effects.

    \item \textbf{Explicit Style Distance Constraint:} We propose an explicit style-distance constraint to robustly manage style-induced intra-class variations, enhancing model robustness against non-categorical variations.

    \item \textbf{Explicit Orthogonality Constraints:} We introduce rigorous orthogonality constraints to strictly decouple common and style subspaces, achieving thorough feature disentanglement.

    \item \textbf{Significant Generalization Improvement Across Datasets and Architectures:} Comprehensive experiments demonstrate that OSCS-SupCon consistently outperforms state-of-the-art supervised contrastive methods across multiple datasets and architectures, particularly excelling in fine-grained classification scenarios.
\end{itemize}

The remainder of this paper is organized as follows. Section~\ref{sec2} reviews related contrastive learning methods and recent advances. Section~\ref{sec3} briefly introduces foundational concepts. Section~\ref{sec4} presents our proposed OSCS-SupCon method in detail. Extensive experimental results and analyses are provided in Section~\ref{sec5}. Finally, Section~\ref{sec6} concludes the paper and discusses directions for future research.

\section{Related Work}
\label{sec2}

\subsection{Contrastive Methods}

Discriminative approaches grounded in contrastive learning within embedding spaces have recently demonstrated substantial potential, achieving state-of-the-art performance~\cite{LIU2023, HU2024}. Wu et al.~\cite{WU2018} introduced a versatile metric loss known as Include and Exclude (IE) loss, effectively enhancing inter-class feature separation and intra-class compactness, particularly beneficial in scenarios with small batch sizes without complex sampling strategies. Sohn~\cite{Sohn2016} extended the traditional triplet loss by proposing an $(N+1)$-tuplet loss, forming the basis of efficient and scalable multi-class $N$-pair loss, reducing dependency on hard negative mining. Wang et al.~\cite{Wang2019} further presented the multi-similarity (MS) loss involving iterative pair sampling and weighting, effectively capturing informative sample pairs to boost recognition performance. These early methods laid important groundwork, yet they did not explicitly disentangle category-relevant (common) and irrelevant (style) features, a challenge explicitly addressed by our proposed OSCS-SupCon.

\subsection{Self-Supervision Based Contrastive Methods}

Self-supervised learning has gained traction as an effective paradigm for representation learning through pretext tasks distinct from the primary supervised objective~\cite{Cheng2021,Chen2021,Henaff2020,ZHAO2024}. Notably, SimCLR~\cite{Chen2020} generates positive pairs via data augmentation and employs a nonlinear projection head to improve visual representations without explicit labels. BYOL~\cite{Grill2020} deviates from traditional contrastive learning paradigms by eliminating negative samples entirely, leveraging two neural networks predicting each other's representations to achieve robust performance against diverse augmentations. Despite their advancements, these methods rely heavily on augmentation strategies and lack explicit mechanisms to manage intra-class variations, particularly the disentanglement of category-irrelevant features—limitations our work explicitly targets.

\subsection{Supervised Contrastive Learning and Variants}

Supervised Contrastive Learning (SupCon)~\cite{khosla2020supervised} extends the self-supervised SimCLR framework by leveraging label information within a two-stage learning paradigm. Initially, an InfoNCE-based contrastive loss is applied to multiple augmented views per image to enhance intra-class compactness and inter-class separability. Subsequently, the learned encoder is frozen, and a classifier is trained using the resulting discriminative embeddings. However, the standard SupCon loss simultaneously compares each anchor against multiple negative samples, inherently diluting the impact of crucial negative pairs and neglecting explicit disentanglement of common and style features.

To reduce computational overhead from extensive multiview augmentations, SelfCon~\cite{baeSelfContrastiveLearningSingleviewed2022} simplifies augmentation by employing a single augmented view per image and introducing an auxiliary sub-network for dual representation generation. However, SelfCon retains the unified InfoNCE-based embedding space, encountering negative sample dilution and lacking explicit mechanisms for feature disentanglement.

Recently, CS-SupCon~\cite{dornaika2025deep} explicitly addressed feature-space disentanglement by partitioning the embedding space into distinct common and style subspaces and introducing explicit style contrastive terms. Although promising, CS-SupCon still relies on the InfoNCE-based contrastive loss, inherently encountering negative sample dilution effects and lacking rigorous orthogonality constraints, leading to partial coupling between subspaces.

Other recent studies have explored supervised contrastive learning from different angles. Xue et al.~\cite{xue2023features} analyzed simplicity bias, revealing its influence on class collapse without explicitly targeting feature disentanglement or negative sample dilution. Azizi et al.~\cite{azizi2023robust} proposed a robust, data-efficient self-supervised approach enhancing generalization in medical imaging, though still neglecting explicit feature subspace disentanglement. Zhang et al.~\cite{zhang2024hierarchical} combined multi-label contrastive learning with KNN for hierarchical text classification, similarly overlooking explicit disentanglement. Duan et al.~\cite{DUAN2025} introduced a unified deep contrastive framework for supervised regression and classification tasks but did not explicitly address negative sample dilution or subspace orthogonality.

In contrast, our proposed OSCS-SupCon explicitly addresses these critical limitations by integrating three distinct enhancements: (i) a novel sigmoid-based pairwise contrastive loss with learnable parameters (\(t, b\)) effectively mitigating negative sample dilution, (ii) an explicit style-distance constraint robustly managing style-induced intra-class variations, and (iii) rigorous orthogonality constraints explicitly imposed between common and style feature subspaces. These innovations significantly enhance discriminative power, feature disentanglement, and robustness, directly overcoming the shortcomings identified in existing supervised contrastive frameworks.

\section{Preliminaries}
\label{sec3}

In this section, we briefly review the core components and training procedure of the recently proposed CS-SupCon framework~\cite{dornaika2025deep}, which forms the foundational basis motivating our OSCS-SupCon method.

\subsection{Overview of CS-SupCon}

Similar to the standard supervised contrastive learning (SupCon) framework~\cite{khosla2020supervised}, CS-SupCon comprises three essential modules: (i) an augmentation module generating multiple augmented views for each input image, (ii) an encoder extracting deep features from the augmented views, and (iii) a projection head mapping these deep features into a normalized embedding space suited for contrastive metric learning.

A crucial innovation of CS-SupCon lies in the explicit partitioning of the embedding space into two distinct subspaces: a \emph{common subspace} $\cvect \in \mathbb{R}^{D_c}$ capturing class-relevant (category) information, and a \emph{style subspace} $\svect \in \mathbb{R}^{D_s}$ representing class-irrelevant (style-related) variations. Formally, the final embedding $\zvect$ is expressed as:
\begin{equation}
    \zvect = [\cvect; \svect], \quad D_p = D_c + D_s.
\end{equation}

\begin{figure}[!htbp]
    \centering
    \includegraphics[width=\textwidth]{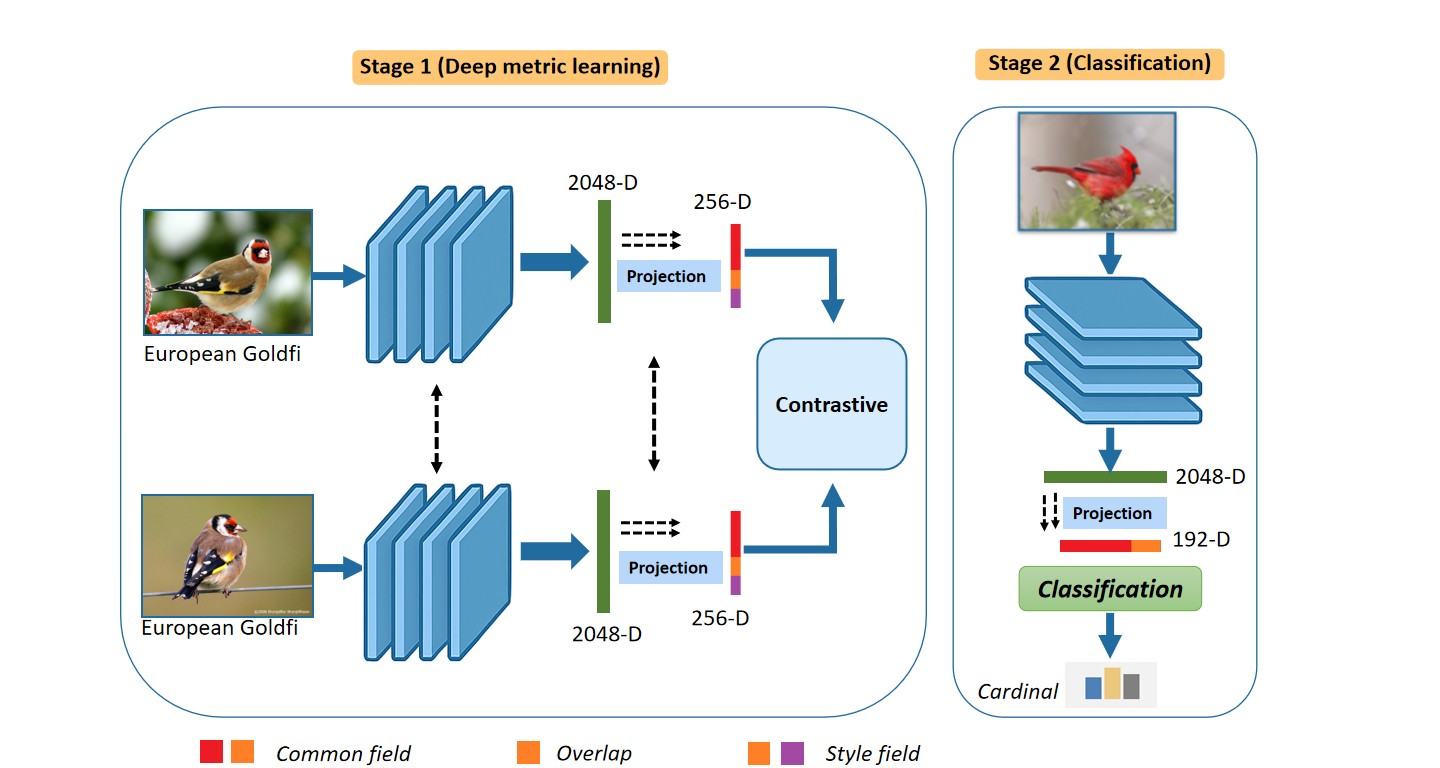}
    \caption{CS-SupCon. Stage 1: Deep metric learning explicitly partitions embeddings into common and style subspaces. Stage 2: A linear classifier is trained exclusively on common features, excluding style-related variations.}
    \label{fig:CS-SCL}
\end{figure}

The complete two-stage training strategy of CS-SupCon explicitly partitions embeddings into common and style subspaces, as illustrated in Figure~\ref{fig:CS-SCL}.

Additionally, CS-SupCon introduces an optional \emph{overlapping mechanism}, allowing specific dimensions to be shared simultaneously by common and style subspaces. This flexible overlapping strategy enhances adaptability to diverse data distributions and empirically improves model generalization performance~\cite{dornaika2025deep}. However, in this work, we build our proposed OSCS-SupCon upon the non-overlapping variant of CS-SupCon, explicitly employing a sigmoid-based contrastive loss to enhance negative pair distinction, while explicitly enforcing orthogonality between common and style features to ensure thorough disentanglement.

\subsection{CS-SupCon Loss and Training Procedure}

Let $I$ represent the set of indices of augmented samples within a training batch. For each sample index $i \in I$, we denote $P(i)$ as the set of indices corresponding to positive samples sharing the same class label as sample $i$. CS-SupCon jointly optimizes the encoder and projection head by minimizing the following composite loss function:
\begin{equation} 
\label{eq:cs_supcon_loss}
\begin{split}
\mathcal{L}_{\text{CS-SupCon}} = 
\sum_{i \in I} \frac{1}{\vert P(i) \vert } \sum_{p \in P(i)} 
\bigg\{ 
& - \log \frac{\exp(\cvect_i \cdot \cvect_p / \tau)}{\sum_{j \in I\setminus \{i\}} \exp(\cvect_i \cdot \cvect_j / \tau)} \\
& + \alpha \log \frac{\exp(\svect_i \cdot \svect_p / \tau)}{\sum_{j \in I\setminus \{i\}} \exp(\svect_i \cdot \svect_j / \tau)} 
- \beta \| \svect_i - \svect_p \| 
\bigg\}.
\end{split}
\end{equation}  

In Eq.~\eqref{eq:cs_supcon_loss}, the temperature parameter $\tau$ controls the similarity scale, and hyperparameters $\alpha$ and $\beta$ balance contributions of style terms. Specifically, this loss consists of three components:

\begin{itemize}
    \item The first term, similar to SupCon, promotes compactness among common features of positive pairs.
    \item The second term explicitly encourages diversity among style features of positive pairs, enhancing style variability.
    \item The third term imposes a Euclidean penalty, explicitly enforcing distances between style representations of positive pairs to reinforce style-content disentanglement.
\end{itemize}

\subsection{Classifier Training on Common Features}

After optimizing the encoder and projection head using Eq.~\eqref{eq:cs_supcon_loss}, their parameters are frozen. Subsequently, a linear classifier is trained exclusively using the common features $\cvect$. Given an input sample represented by common features $\cvect$, the classifier predicts class probabilities $\hat{\yvect}\in\mathbb{R}^{C}$, where $C$ denotes the total number of classes. Classifier parameters are optimized by minimizing the cross-entropy loss:
\begin{equation}
\mathcal{L}_{\text{CE}}(\hat{\yvect}, \yvect) = -\sum_{k=1}^{C} y_k \log(\hat{y}_k),
\end{equation}
where $\yvect\in\mathbb{R}^{C}$ is the ground-truth one-hot label, and $\hat{y}_k$ is the predicted probability of class $k$. This exclusive training on common features ensures the classifier explicitly leverages only discriminative, category-specific information, thereby enhancing generalization by reducing interference from intra-class style variations.

\section{Proposed Approach: Orthogonal Sigmoid-based Common and Style Supervised Contrastive Learning (OSCS-SupCon)}
\label{sec4}

In this section, we formally introduce our proposed Orthogonal Sigmoid-based Common and Style Supervised Contrastive Learning (OSCS-SupCon) framework. Our method explicitly employs a sigmoid-based contrastive loss to enhance negative sample distinction, incorporates an explicit style-distance constraint, and strictly enforces orthogonality between common and style features to achieve thorough feature disentanglement.

\subsection{Motivation and Overview}

Although existing methods explicitly partition embeddings into common and style subspaces, their reliance on traditional InfoNCE-based contrastive loss inherently dilutes the contribution of negative samples. Furthermore, inadequate explicit constraints between these subspaces lead to residual coupling, severely limiting robustness and generalization, particularly in fine-grained classification tasks. To effectively address these limitations, we propose OSCS-SupCon, explicitly integrating a sigmoid-based pairwise contrastive loss and an explicit style-distance constraint, along with rigorous orthogonality constraints between common and style subspaces, thereby significantly enhancing feature disentanglement and robustness.

\begin{figure}[!htbp]
    \centering
    \includegraphics[width=\textwidth]{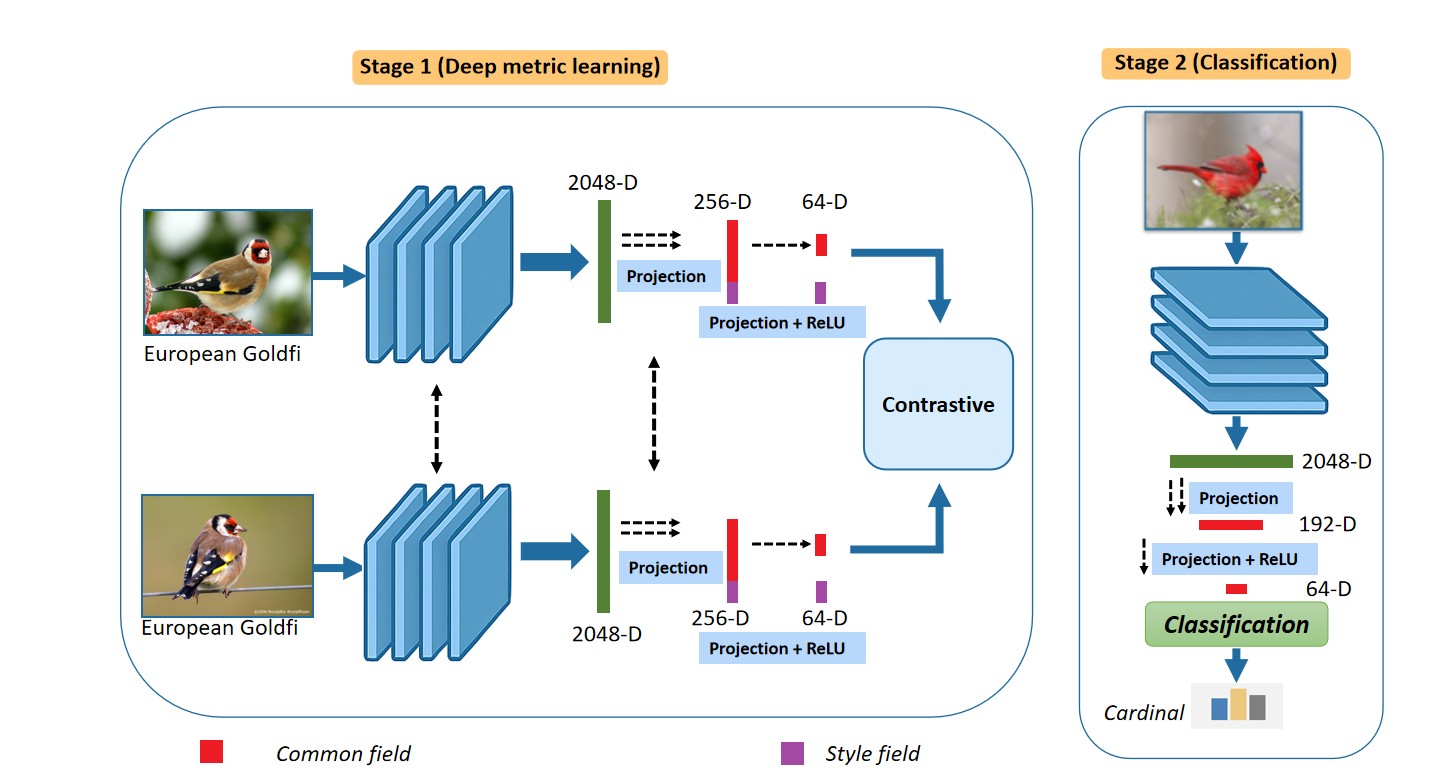}
    \caption{Orthogonal Sigmoid-based Common and Style Supervised Contrastive Learning (OSCS-SupCon). Stage 1: Deep metric learning explicitly partitions embeddings into common and style subspaces, employs a sigmoid-based contrastive loss with learnable parameters, explicitly imposes a style-distance penalty, and enforces orthogonality between common and style features. Stage 2: A linear classifier is trained exclusively on orthogonally disentangled common features.}
    \label{fig:OSCS-SCL}
\end{figure}

The overall two-stage training procedure of OSCS-SupCon is illustrated in Figure~\ref{fig:OSCS-SCL}, clearly highlighting the critical innovations compared to previous methods.

\subsection{Proposed OSCS-SupCon Loss Function}

Formally, we define the proposed OSCS-SupCon loss as:
\begin{eqnarray}
\label{eq:oscs_supcon_loss_final}
\mathcal{L}_{\text{OSCS-SupCon}} &=& 
-\frac{1}{ \vert I  \vert ^2}\sum_{u\in I}\sum_{v\in I}\log\frac{1}{1 + e^{z_{uv}(- t \,  \cvect_u \cdot \cvect_v + b)}} \nonumber \\
& &\quad - \beta\frac{1}{\vert I \vert }\sum_{i\in I}\frac{1}{ \vert P(i) \vert }\sum_{p\in P(i)}\|\svect_i-\svect_p\|  \\
& &\quad + \gamma \frac{1}{\vert I \vert }\sum_{i\in I} \vert \cos (O_i) \vert , \nonumber
\end{eqnarray}

In the above equation, indices $u$ and $v$ denote an arbitrary pair of samples within the current training batch, explicitly covering all possible positive and negative pairs. Here, $O_i$ denotes the angle between the style  vector and the projected common vector associated with the sample $i$. Unlike conventional supervised contrastive loss functions that focus mainly on anchor-positive pairs, our sigmoid-based loss explicitly models relationships between all pairs of samples. This comprehensive modeling significantly enhances the model's ability to distinguish subtle differences between positive and negative pairs, thereby improving robustness and generalization. Specifically, the temperature parameter is parameterized by a learnable intermediate parameter $t'$ as $t=\exp(t')$, ensuring positivity and numerical stability during training. The indicator variable $z_{uv}$ equals $+1$ for positive pairs and $-1$ for negative pairs.

The learnable bias term $b$ provides flexible decision boundaries for distinguishing positive and negative pairs. Specifically, a positive $b$ biases the model toward classifying pairs as positives, while a negative $b$ increases the classification threshold. The hyperparameter $\beta$ controls the strength of the style-distance penalty, enforcing effective disentanglement by ensuring adequate variability among style representations of samples from the same class.

Additionally, OSCS-SupCon explicitly introduces a novel orthogonality constraint term controlled by hyperparameter $\gamma$. This orthogonality constraint explicitly enforces the common and style subspaces to be strictly orthogonal by minimizing the absolute cosine similarity between their corresponding feature vectors:
\begin{equation}
\vert   \text{Cos}(O_i) \vert  = \frac{ \vert  \cvect'_i \cdot \svect_i \vert }{\vert \vert  \cvect'_i\vert \vert \:   \vert \vert  \svect_i \vert \vert },
\end{equation}
where $\cvect'_i$ represents the transformed common feature vector after an additional linear projection with ReLU activation, and $\svect_i$ denotes the corresponding style feature vector for the $i$-th sample. Specifically, $O_i$ denotes the angle between these two vectors, and thus $\vert {Cos}(O_i) \vert $ explicitly measures their absolute cosine similarity. Minimizing this absolute cosine similarity explicitly ensures strict orthogonality between common and style subspaces, further enhancing feature disentanglement, interpretability, and robustness.

In summary, by integrating the proposed sigmoid-based loss with an explicit orthogonality constraint, and incorporating the style-distance penalty introduced in prior work, OSCS-SupCon achieves enhanced discriminative power and effective feature disentanglement compared to previous methods.

\section{Performance evaluation}
\label{sec5}
\subsection{Datasets, Encoders and Methods}

We evaluate the proposed OSCS-SupCon method on the following public image datasets: CIFAR10, CIFAR100, Tiny-ImageNet, CUB200-2011, Stanford Dogs, and PASCAL VOC 2005. We adopt standard training/testing splits unless stated otherwise. Table~\ref{tab:datasets} reports the details of these datasets.

\begin{table}[htbp]
\caption{Datasets used in the experiments.} 
\label{tab:datasets}
\centering
  \begingroup
    \color{black}
\resizebox{0.8\textwidth}{!}{
\begin{tabular}{l  m{1.8cm} r  r  r}
\hline
Dataset & Image size & $ \sharp$ classes & \multicolumn{2}{c}{Standard Split} \\
\cline{4-5}
& & & Training Set & Test Set \\
\hline
CIFAR10 & 32 $\times$ 32 & 10 & 50,000 & 10,000 \\
CIFAR100 & 32  $\times$ 32 & 100 & 50,000 & 10,000 \\
Tiny-ImageNet & 64 $\times$ 64 & 200 & 100,000 & 10,000 \\
CUB200-2011 & 224 $\times$ 224 & 200 & 5,994 & 5,794 \\
Stanford Dogs & 224 $\times$  224 & 120 & 12,000 & 8,580 \\
PASCAL VOC 2005 & 224  $\times$ 224 & 4 & 1,843 & 389 \\
\hline
\end{tabular}
}
\endgroup
\end{table}

Various backbone encoders are used. We use the pretrained ResNet18~\cite{He15_resnet50}, ResNet50~\cite{He15_resnet50}, and ConvNeXt-T~\cite{liu2022convnet} as CNN encoders, along with two pretrained Transformers from the Model Zoo library: TinyViT (5M) and TinyViT (21M). The evaluation includes the following methods: 
(1) a baseline using the backbone encoder with a default classification head, 
(2) SimCLR~\cite{Chen2020}, an unsupervised contrastive learning approach, 
(3) BYOL~\cite{Grill2020}, a self-supervised learning method, 
(4) SupCon~\cite{khosla2020supervised}, a supervised contrastive learning technique,  
(5) SelfCon~\cite{baeSelfContrastiveLearningSingleviewed2022}, a single-view supervised contrastive learning framework utilizing a sub-network to reduce the need for multiview augmentation,  
(6) the previously proposed CS-SupCon~\cite{dornaika2025deep}, which explicitly partitions features into common and style subspaces,  
(7) CS-SupCon with overlap~\cite{dornaika2025deep}, which introduces shared dimensions between common and style subspaces, and  
(8) the proposed OSCS-SupCon, which employs a sigmoid-based contrastive loss with explicit style-distance and orthogonality constraints.

For all encoders and methods, we use a projection head composed of a linear layer projecting features to a 256-dimensional space, i.e., $D_p = 256$. The previously proposed CS-SupCon~\cite{dornaika2025deep} partitions features into common ($D_c = 192$) and style ($D_s = 64$) subspaces. In our proposed OSCS-SupCon, we retain these dimensional settings and replace the traditional contrastive loss with our sigmoid-based loss function to enhance feature discrimination. Moreover, we explicitly maintain the style-distance penalty from CS-SupCon to enforce style variation constraints. Crucially, OSCS-SupCon introduces an additional linear projection with a ReLU activation, transforming common features into a $64$-dimensional space aligned with the style features, and explicitly imposes orthogonality constraints between these two subspaces, thereby achieving rigorous feature disentanglement and improved robustness.

\subsection{Implementation Details}

We apply standard data augmentations, including scaling, cropping, and horizontal flipping. Optimization is performed using stochastic gradient descent (SGD) with a momentum of 0.9, weight decay of $1\times 10^{-4}$, and cosine annealing learning rate scheduling. The batch size and learning rates used are consistent with prior experiments (Table~\ref{tab:hyperparam}). 
Specifically, the learnable temperature parameter $t$ (parameterized via $t'$ as $t = e^{t'}$) and bias $b$ are initialized as $t=0.1$ and $b=0$, respectively, and optimized jointly with model parameters.

\begin{table*}[h!]
  \begingroup
    \color{black}
\centering
\caption{Value of the main hyperparameters used by the proposed OSCS-SupCon method. S1: Stage 1, S2: Stage 2, LR: Initial learning rate, Ep: number of epochs, BS: Batch size (S1/S2). These hyperparameters are selected based on previous empirical experiments, ensuring effective and stable training convergence across datasets.}
\label{tab:hyperparam}
\resizebox{0.9\textwidth}{!}{
\begin{tabular}{p{4cm} p{1.5cm} p{1.5cm} p{1.5cm} p{1.5cm} p{1.5cm}}
\hline
Dataset & LR in S1 & Ep. in S1 & LR in S2 & Ep. in S2 & BS \\ \hline
& & & & & \\  
 CIFAR10 & 0.5 & 1000 & 3 & 100 & 1024/512 \\
 CIFAR100 & 0.5 & 1000 & 3 & 100 & 1024/512 \\
 Tiny-ImageNet & 0.5 & 1000 & 3 & 100 & 1024/512\\
 CUB200-2011 & 0.1 & 1000 & 0.1 & 100 & 96/48\\
 Stanford Dogs (TinyViT(5M)) & 0.1 & 1000 & 0.1 & 100 & 96/48\\
 Stanford Dogs (ConvNext-tiny) & 0.1 & 1000 & 0.1 & 100 & 96/48\\
 PASCAL VOC 2005 & 0.1 & 1000 & 0.1 & 100 & 96/48\\
\hline
\end{tabular}
}
\endgroup
\end{table*}

\subsection{Experimental Results}

Tables~\ref{tab:results} and~\ref{tab:my-table} summarize the classification accuracy of the proposed OSCS-SupCon method compared to existing contrastive learning methods across multiple datasets and backbone encoders. Clearly, OSCS-SupCon consistently achieves state-of-the-art performance, significantly outperforming classical SupCon, SelfCon, and CS-SupCon approaches. Particularly, the introduction of the sigmoid-based contrastive loss effectively mitigates the negative sample dilution issue inherent in traditional methods. Additionally, the explicit orthogonality constraints between common and style subspaces, along with the explicit style-distance penalty, lead to substantial accuracy improvements, especially noticeable on fine-grained datasets such as CUB200-2011 and Stanford Dogs, as well as across both CNN and Transformer architectures.

\begin{table*}[h!]
\centering
\caption{Method comparison using five image datasets and three different encoders.}
\label{tab:results}
\resizebox{1.1\textwidth}{!}{
\begin{tabular}{l|cccccc}
\hline
\backslashbox{Method}{Dataset} & CIFAR10 & CIFAR100 & CUB200-2011 & Stanford Dogs & Stanford Dogs & PASCAL VOC\\
\hline
Backbone & ResNet50 & ResNet50 & TinyViT (21M) & TinyViT (5M) & ConvNeXt-tiny & TinyViT (5M) \\ 
\hline
Baseline & 94.9 & 74.8 & 88.7 & 83.4 & 92.1 & 47.8\\
SimCLR & 93.6 & 73.9 & 80.1 & 77.6 & 86.1 & 47.0 \\
BYOL & 95.0 & 75.0 & 80.7 & 78.3 & 88.2 & 45.8 \\
SupCon & 95.6 & 75.5 & 89.1 & 85.5 & 92.8 & 51.9 \\
SelfCon &95.6 &78.1 &- &- &- &- \\
CS-SupCon & 95.4 & 77.6 & 89.8 & 86.0 & 92.8 & 51.4 \\
CS-SupCon w. ov. & 95.7 & 78.4 & 89.8 & 86.2 & 93.1 & 53.3 \\
OSCS-SupCon (Ours) & \textbf{95.9} & \textbf{79.5} & \textbf{90.5} & \textbf{87.0} & \textbf{94.1} & \textbf{54.3} \\ 
\hline
\end{tabular}}
\end{table*}

\begin{table*}[h!]
\centering
  \begingroup
    \color{black}
\caption{Classification results of different contrastive methods using the ResNet-18 architecture on five datasets.}
\label{tab:my-table}
\resizebox{1.1\textwidth}{!}{
\begin{tabular}{l|ccccc}
\hline
\backslashbox{Method}{Dataset} & CIFAR10 & CIFAR100 & Tiny-ImageNet & CUB200-2011 & Stanford Dogs \\ \hline
Baseline & 94.7 & 72.9 & 57.5 & 55.9 & 61.5 \\
SimCLR & 93.0 & 72.3 & 55.9 & 55.2 & 60.8 \\
BYOL & 94.6 & 73.0 & 57.2 & 56.2 & 62.3 \\
SupCon & 94.7 & 73.6 & 57.7 & 57.1 & 63.0 \\
SelfCon & 95.1 & 74.9 & 59.8 & 60.4 & 65.4 \\
CS-SupCon & 94.9 & 74.1 & 59.0 & 59.1 & 64.8 \\
CS-SupCon w. ov. & 95.2 & 75.2 & 59.9 & 61.0 & 66.3 \\ 
OSCS-SupCon (Ours) & \textbf{95.6} & \textbf{75.8} & \textbf{61.1} & \textbf{62.5} & \textbf{68.2} \\ 
\hline
\end{tabular}
}
\endgroup
\end{table*}

Further rigorous validation through five-fold cross-validation on CIFAR-100 (Table~\ref{tab:cifar100_res50_5fold}) demonstrates that OSCS-SupCon consistently yields statistically significant improvements compared to prior methods. Specifically, OSCS-SupCon achieves an average accuracy improvement of approximately 4.4\% over SupCon, about 1.8\% over SelfCon, and 2.2\% over the original CS-SupCon. A paired Student's t-test ($p<0.05$) confirms the significance of these improvements, with statistically significant results highlighted in bold and marked with a dagger ($^\dagger$). These findings further confirm that explicitly integrating the sigmoid-based pairwise loss, style-distance constraint, and orthogonality constraint significantly enhances the discriminative power and robustness of the learned representations, resulting in improved generalization performance.

\begin{table}[h]
\caption{Five-fold cross-validation Top-1 accuracy (\%) on CIFAR-100 (ResNet-50)}
\label{tab:cifar100_res50_5fold}
\resizebox{\textwidth}{!}{
\begin{tabular}{l|ccccc|c}
\hline
Method & Fold-1 & Fold-2 & Fold-3 & Fold-4 & Fold-5 & Mean \\ \hline
SupCon & 74.8 & 74.6 & 74.5 & 74.7 & 74.4 & 74.6 \\
SelfCon & 76.9 & 77.3 & 77.1 & 77.2 & 77.5 & 77.2 \\
CS-SupCon & 76.6 & 76.9 & 76.7 & 76.8 & 77.0 & 76.8 \\
CS-SupCon w. ov. & 78.0 & 77.7 & 77.9 & 77.8 & 77.6 & 77.8 \\ 
OSCS-SupCon (Ours) & \textbf{79.2}$^\dagger$ & \textbf{79.0}$^\dagger$ & \textbf{78.9}$^\dagger$ & \textbf{78.8}$^\dagger$ & \textbf{79.1}$^\dagger$ & \textbf{79.0}$^\dagger$ \\
\hline
\multicolumn{7}{l}{$^\dagger$: Statistically significant improvement ($p<0.05$) over SupCon, SelfCon, CS-SupCon, and CS-SupCon w. ov. }
\end{tabular}}
\end{table}

\subsection{Learnable Parameters and Sensitivity Analysis}

To comprehensively understand the sensitivity and optimality of hyperparameters and learnable parameters in OSCS-SupCon, we perform detailed analyses on four critical parameters: learnable temperature parameter $t$, bias parameter $b$, hyperparameter $\beta$ (style-distance penalty), and hyperparameter $\gamma$ (orthogonality constraint).

\begin{figure}[htbp]
    \centering
    \begin{tabular}{cc}
       \includegraphics[width=0.5\linewidth]{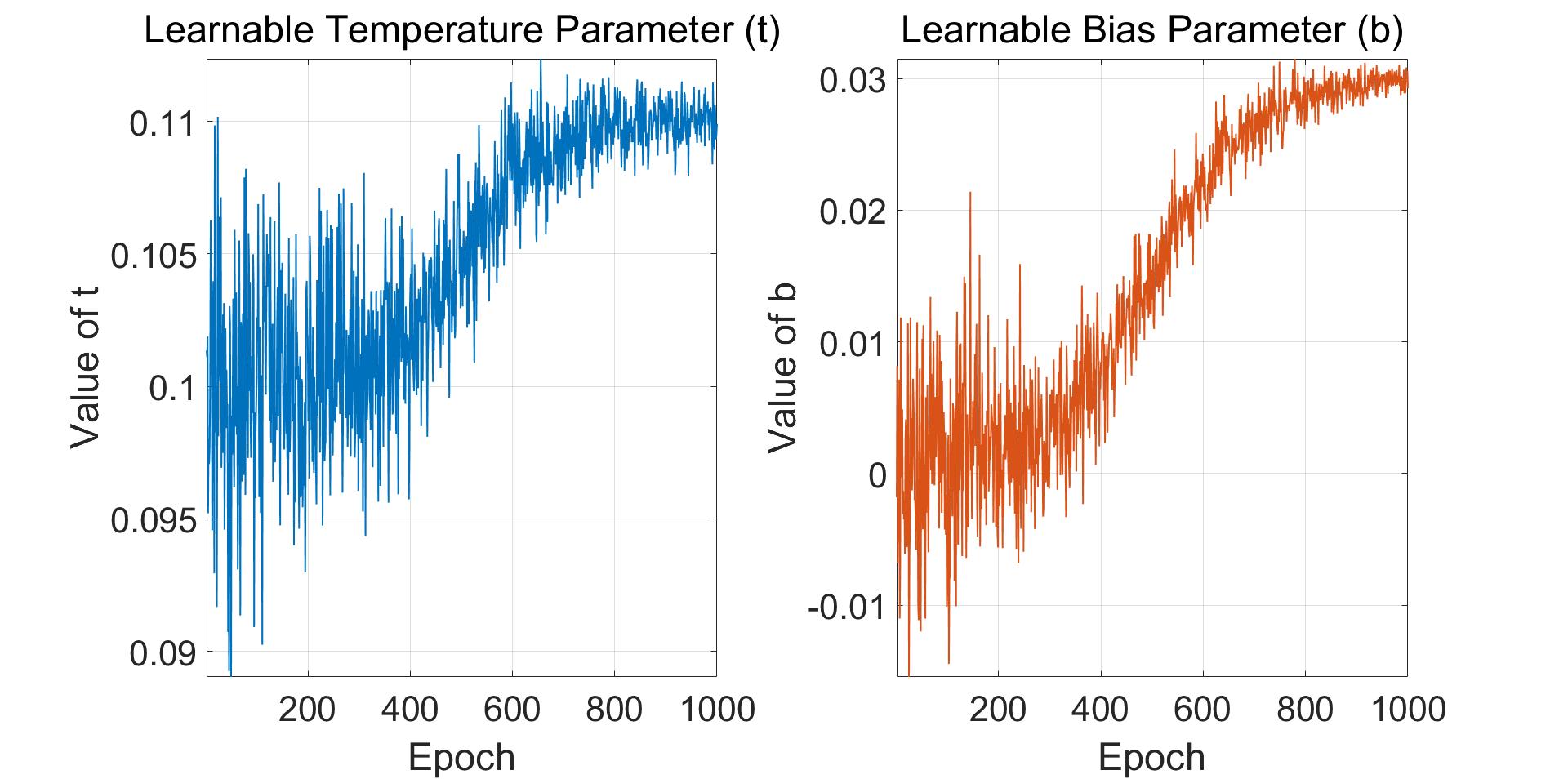} & \includegraphics[width=0.5\linewidth]{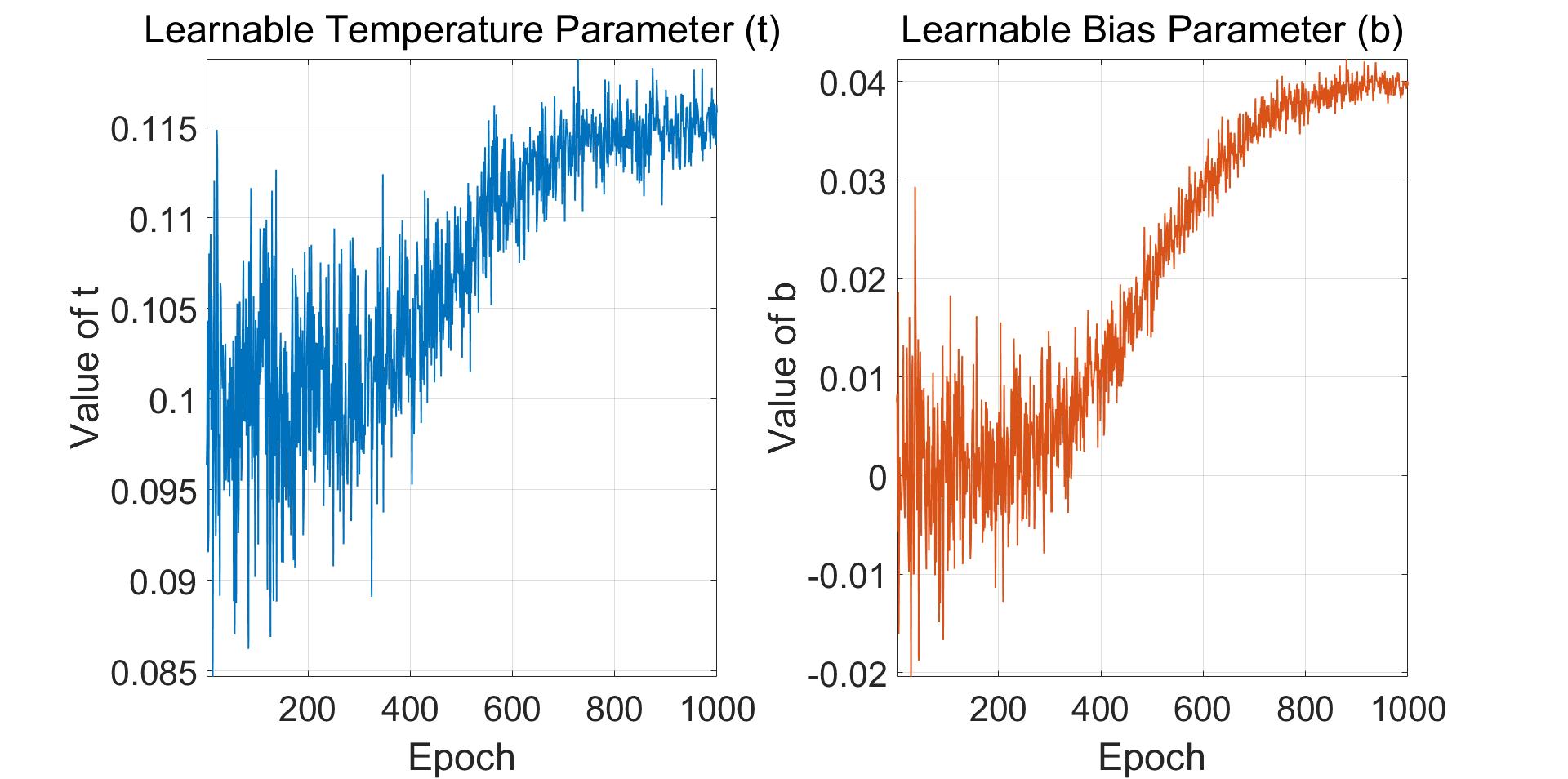}\\
       (a)  CIFAR-10 (ResNet-50) & (b) CIFAR-100 (ResNet-50) \\

 \includegraphics[width=0.5\linewidth]{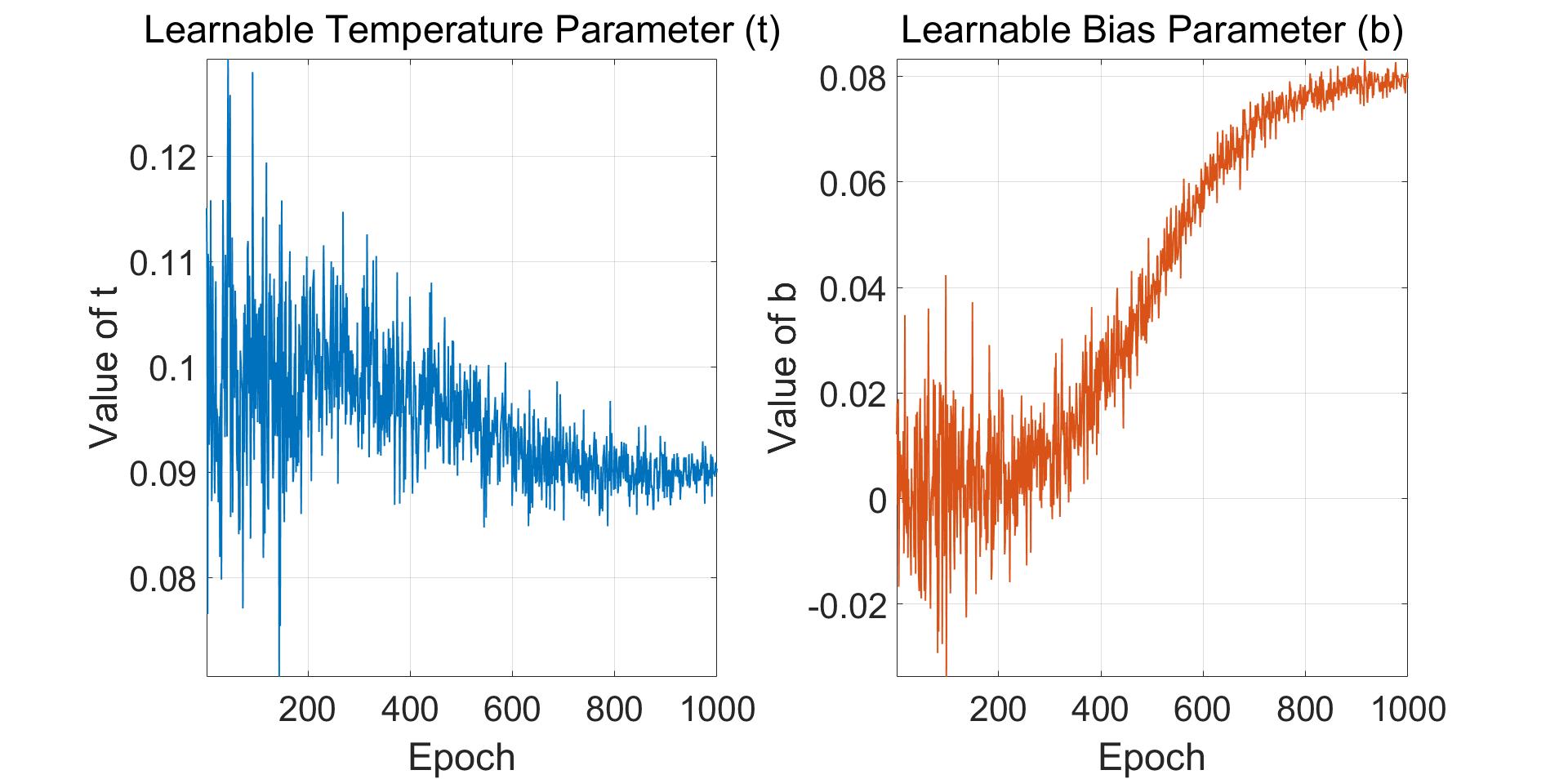} & \includegraphics[width=0.5\linewidth]{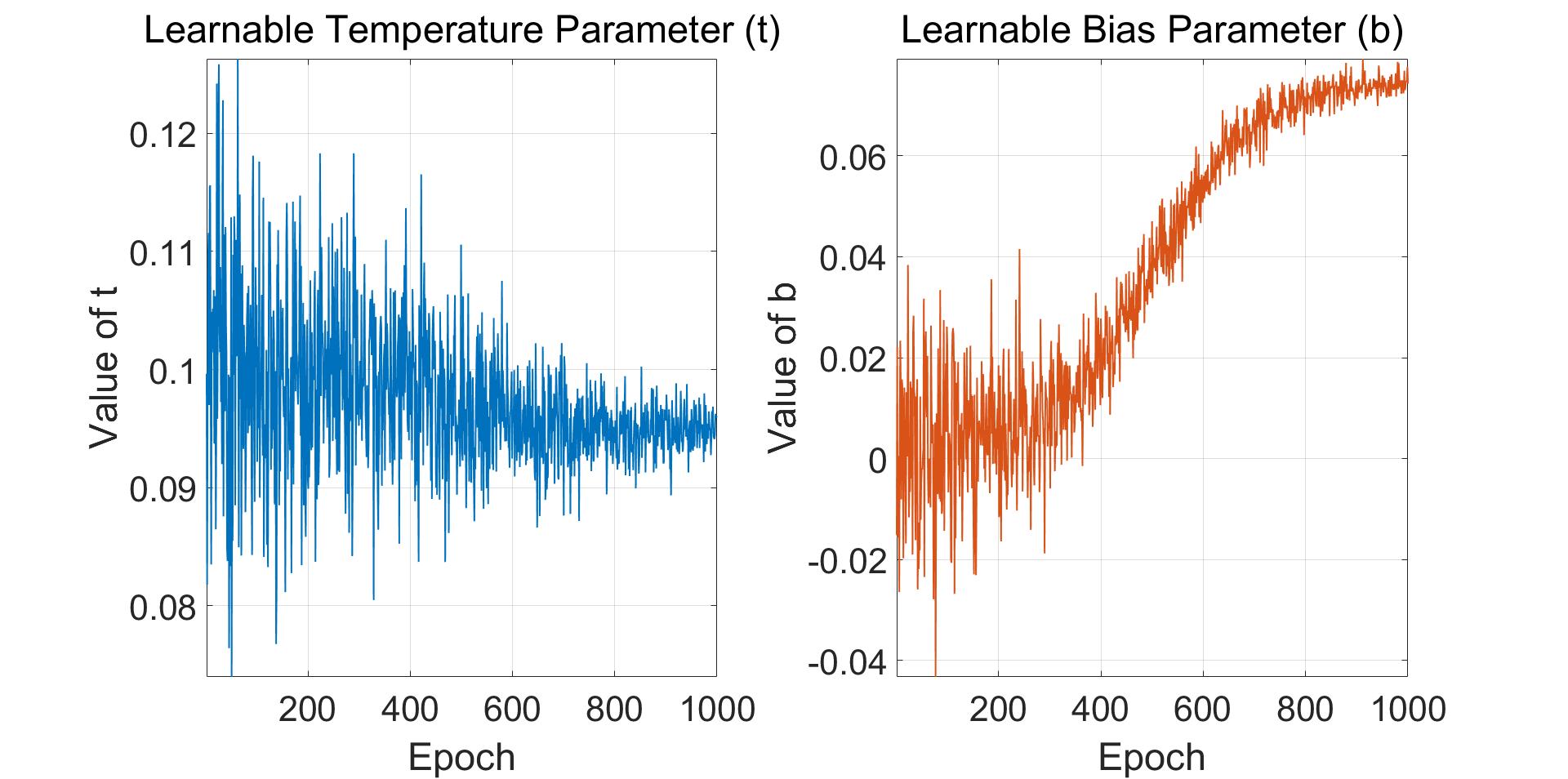}\\
 (c) CUB200-2011 (TinyViT-21M)  & (d) Stanford Dogs (TinyViT-5M)\\
       
        \includegraphics[width=0.5\linewidth]{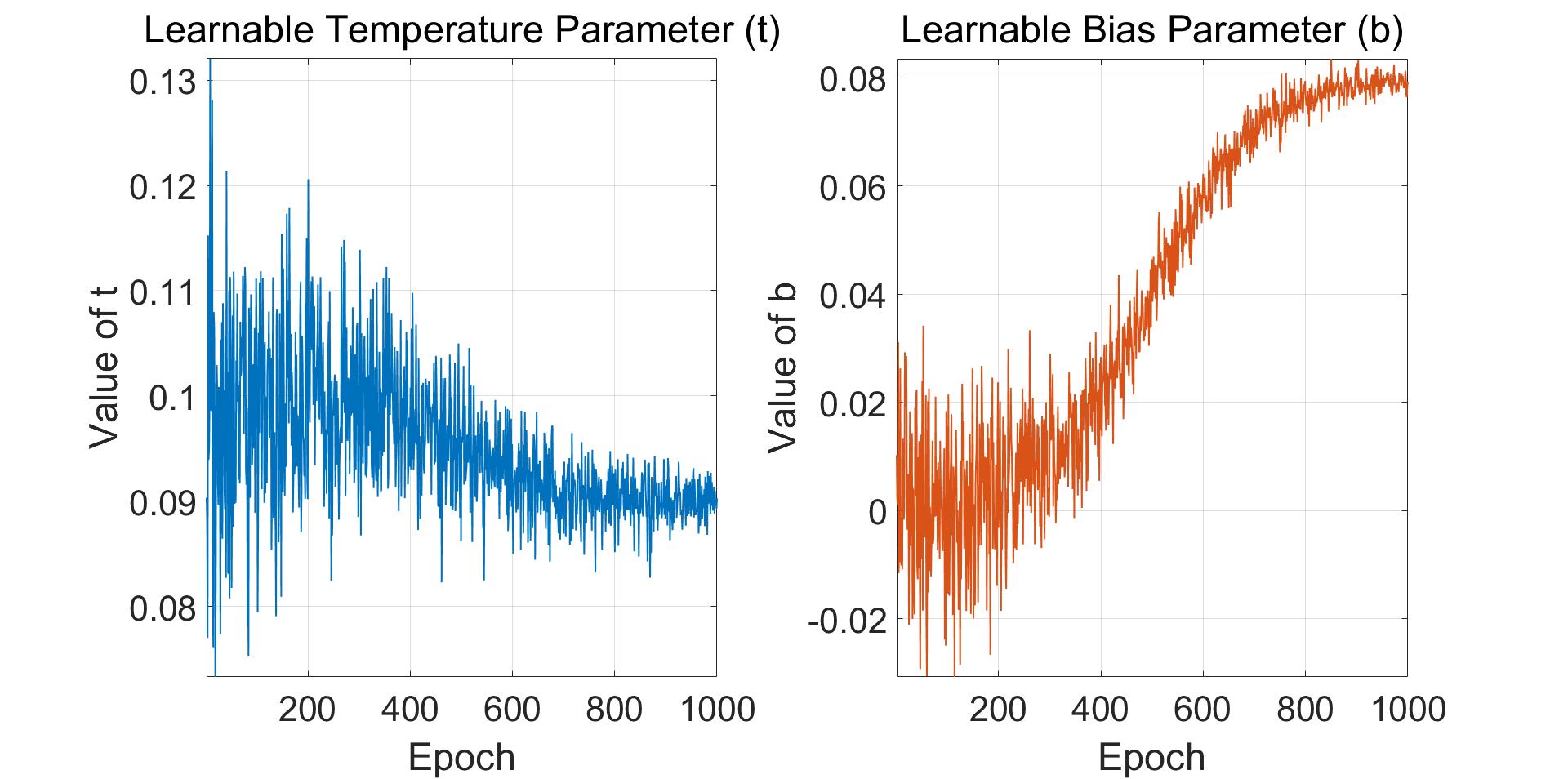} & \includegraphics[width=0.5\linewidth]{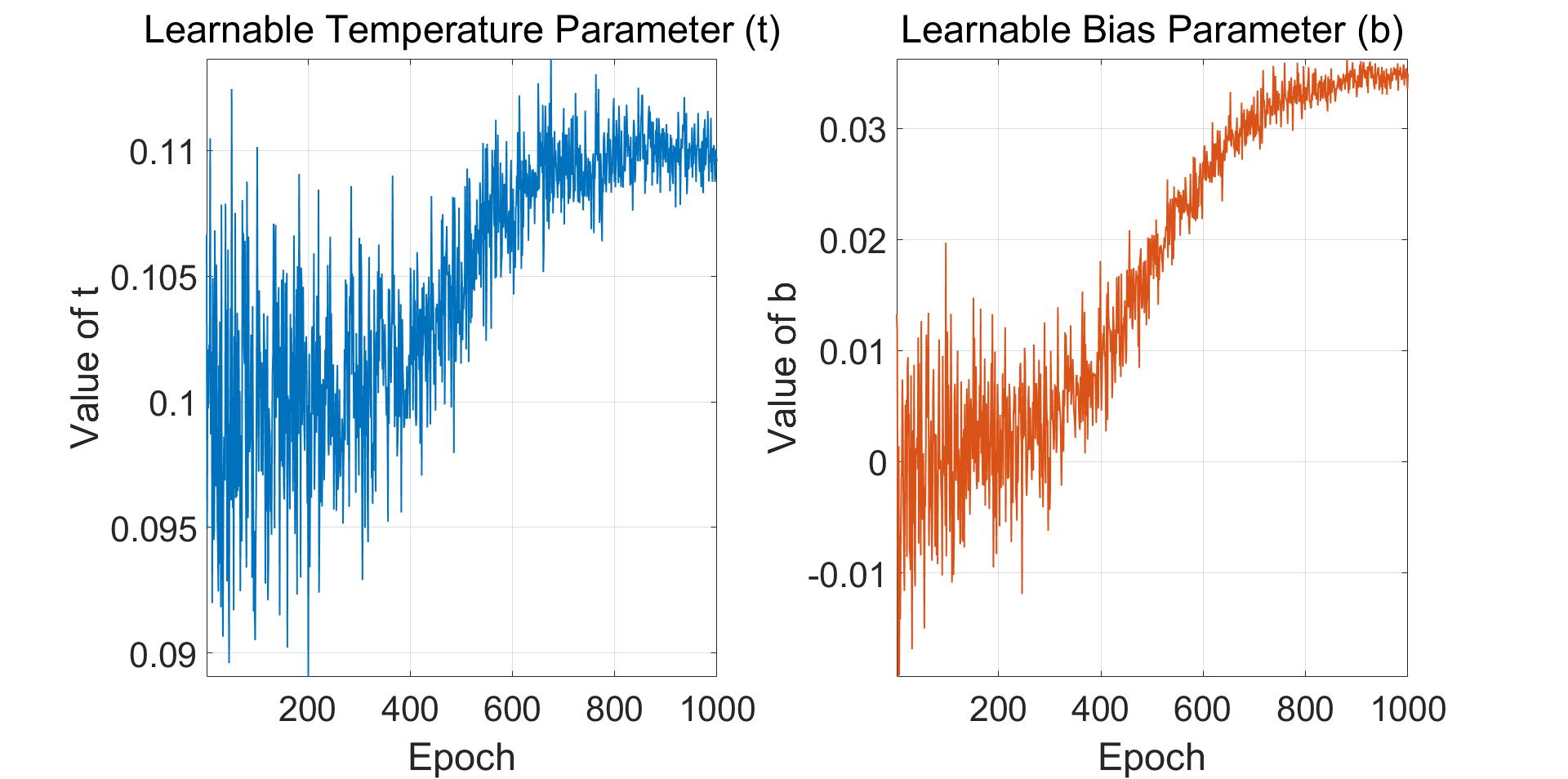} \\
        (e)  Stanford Dogs (ConvNeXt-tiny) &  (f) PASCAL VOC (TinyViT-5M) 
     \end{tabular}
    \caption{Learning curves of temperature and bias parameters across various datasets and backbone encoders. The stable convergence validates the robustness and adaptability of OSCS-SupCon.}
    \label{fig:all_curves}
\end{figure}

\paragraph{Temperature parameter $t$:}
Figure~\ref{fig:all_curves} illustrates the training dynamics of the learnable temperature parameter $t$ across various datasets and backbones. Notably, we observe adaptive convergence behavior of $t$, where general datasets (CIFAR-10, CIFAR-100, PASCAL VOC) stabilize at slightly higher temperatures ($0.112$--$0.118$), promoting smoother decision boundaries suitable for broad inter-class distinctions. In contrast, fine-grained datasets (CUB200-2011, Stanford Dogs) converge at lower temperatures ($0.093$--$0.098$), enabling sharper boundaries to distinguish subtle intra-class differences effectively.

\paragraph{Bias parameter $b$:}
The curves of the learnable bias parameter $b$ (Figure~\ref{fig:all_curves}) demonstrate similarly adaptive behavior. For fine-grained datasets, $b$ stabilizes at slightly higher values (approximately $0.06$--$0.065$), effectively adjusting decision boundaries to capture subtle intra-class variations. Conversely, for general datasets $b$ settles near zero (approximately $0.025$--$0.035$), indicating minimal bias adjustment due to clearer inter-class boundaries.

\begin{figure}[!htbp]
    \centering
    \includegraphics[width=0.85\textwidth]{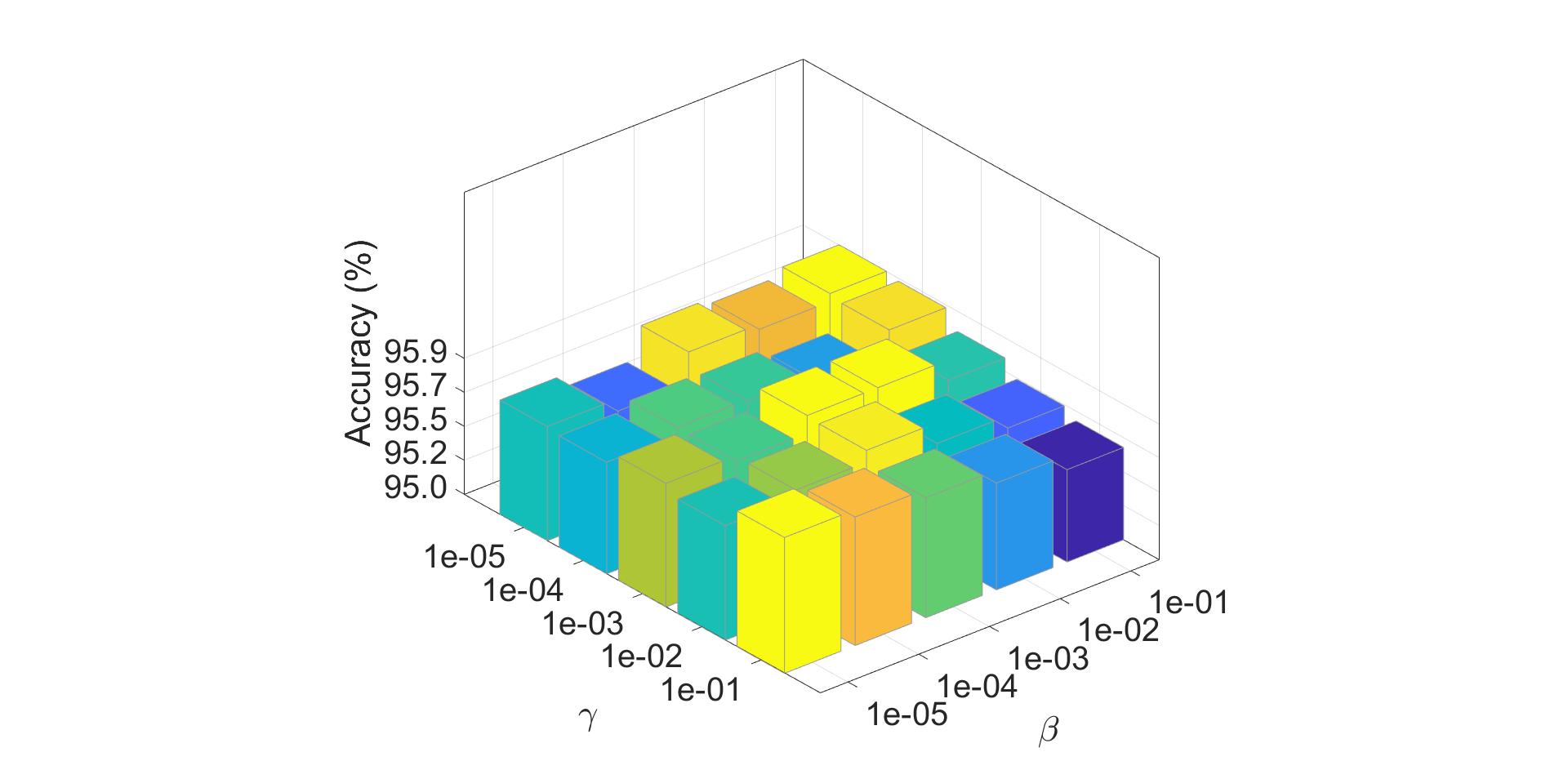}
    \caption{CIFAR-10 (ResNet-50)}
    \label{fig:cifar10_gamma_comparison}
\end{figure}

\begin{figure}[!htbp]
    \centering
    \includegraphics[width=0.85\textwidth]{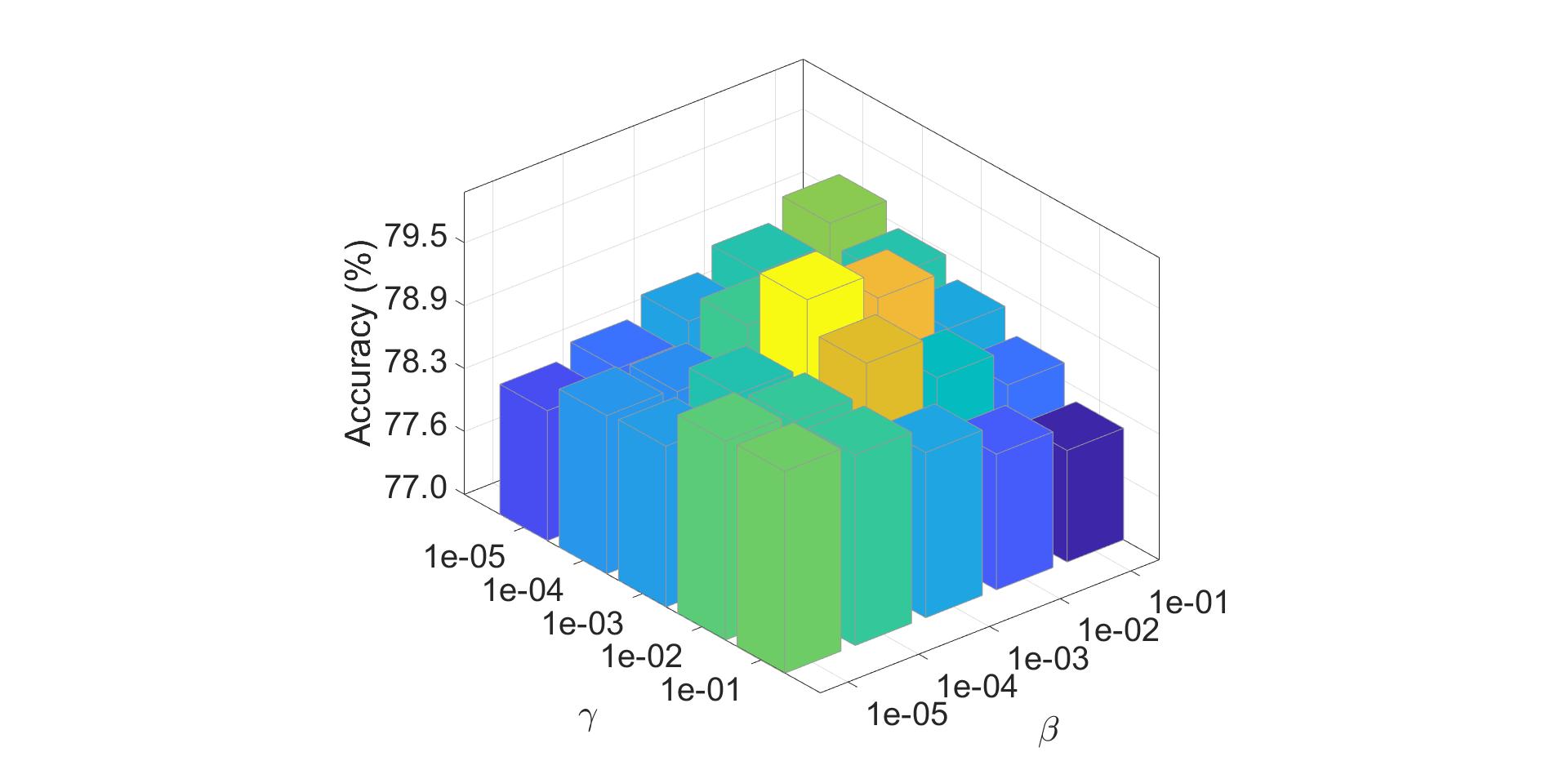}
    \caption{CIFAR-100 (ResNet-50)}
    \label{fig:cifar100_gamma_comparison}
\end{figure}

\begin{figure}[!htbp]
    \centering
    \includegraphics[width=0.85\textwidth]{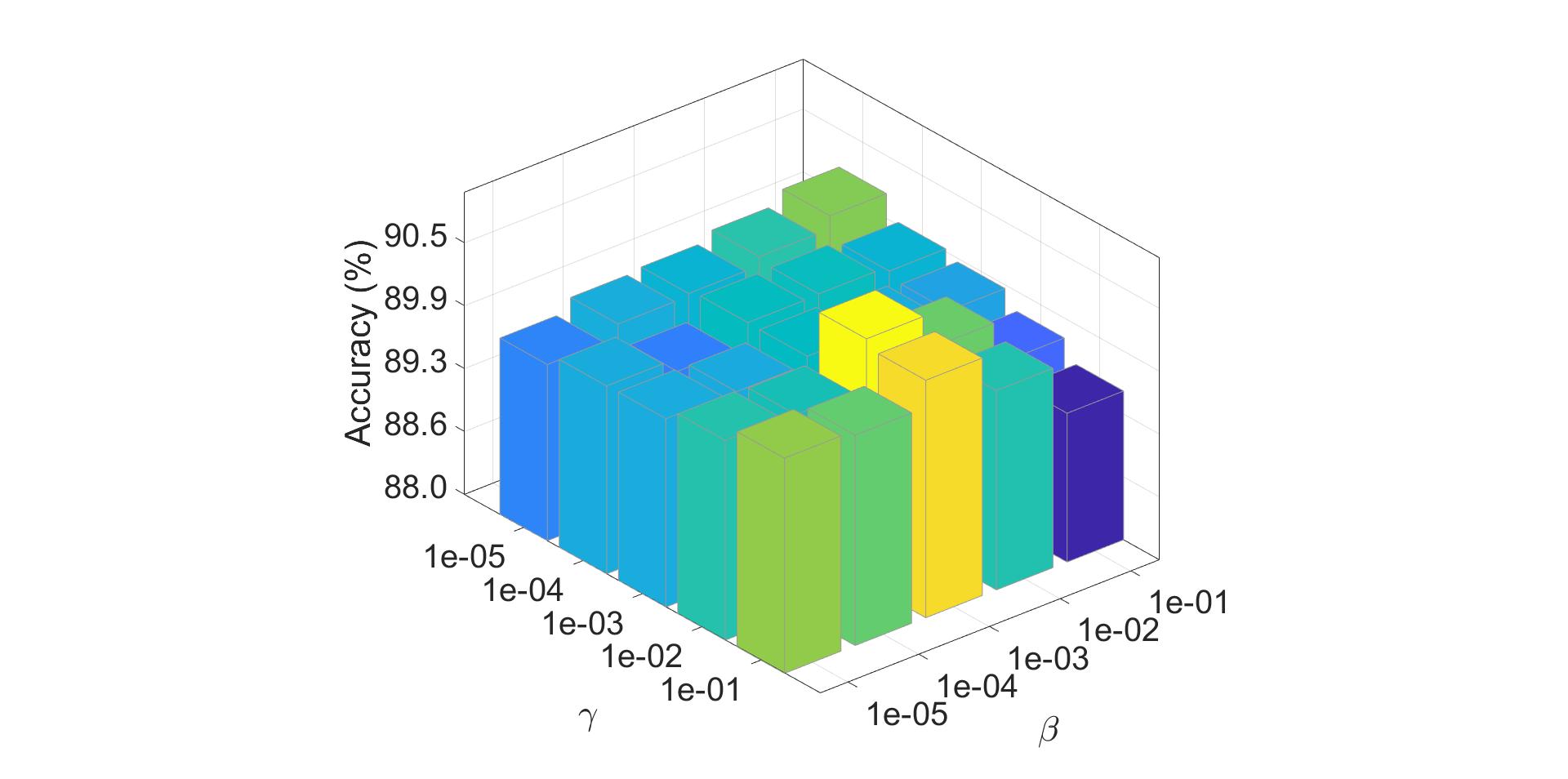}
    \caption{CUB200-2011 (TinyViT-21M)}
    \label{fig:cub200_gamma_comparison}
\end{figure}

\begin{figure}[!htbp]
    \centering
    \includegraphics[width=0.85\textwidth]{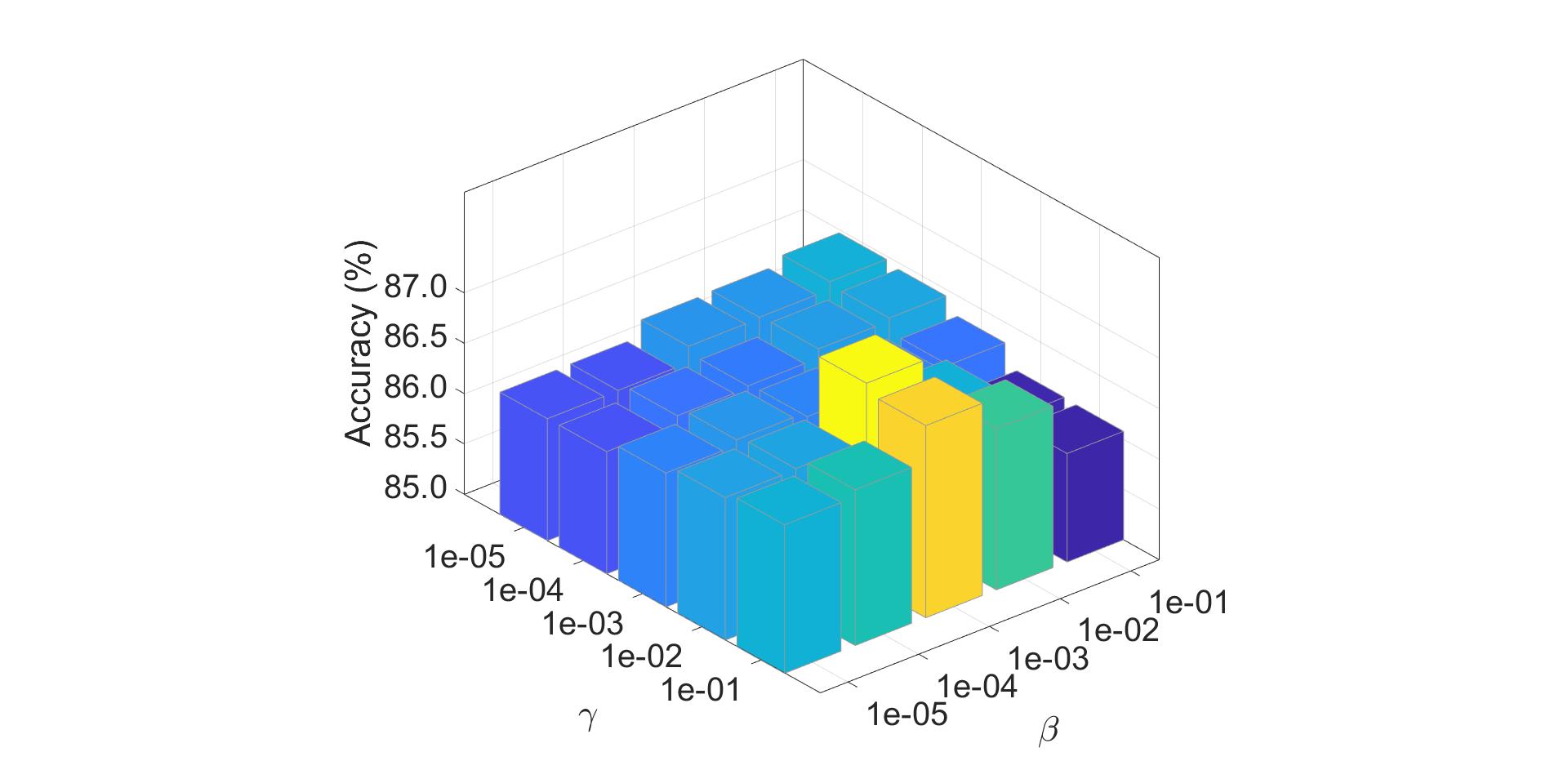}
    \caption{Stanford Dogs (TinyViT-5M)}
    \label{fig:stanforddogs_tinyvit_gamma_comparison}
\end{figure}

\begin{figure}[!htbp]
    \centering
    \includegraphics[width=0.85\textwidth]{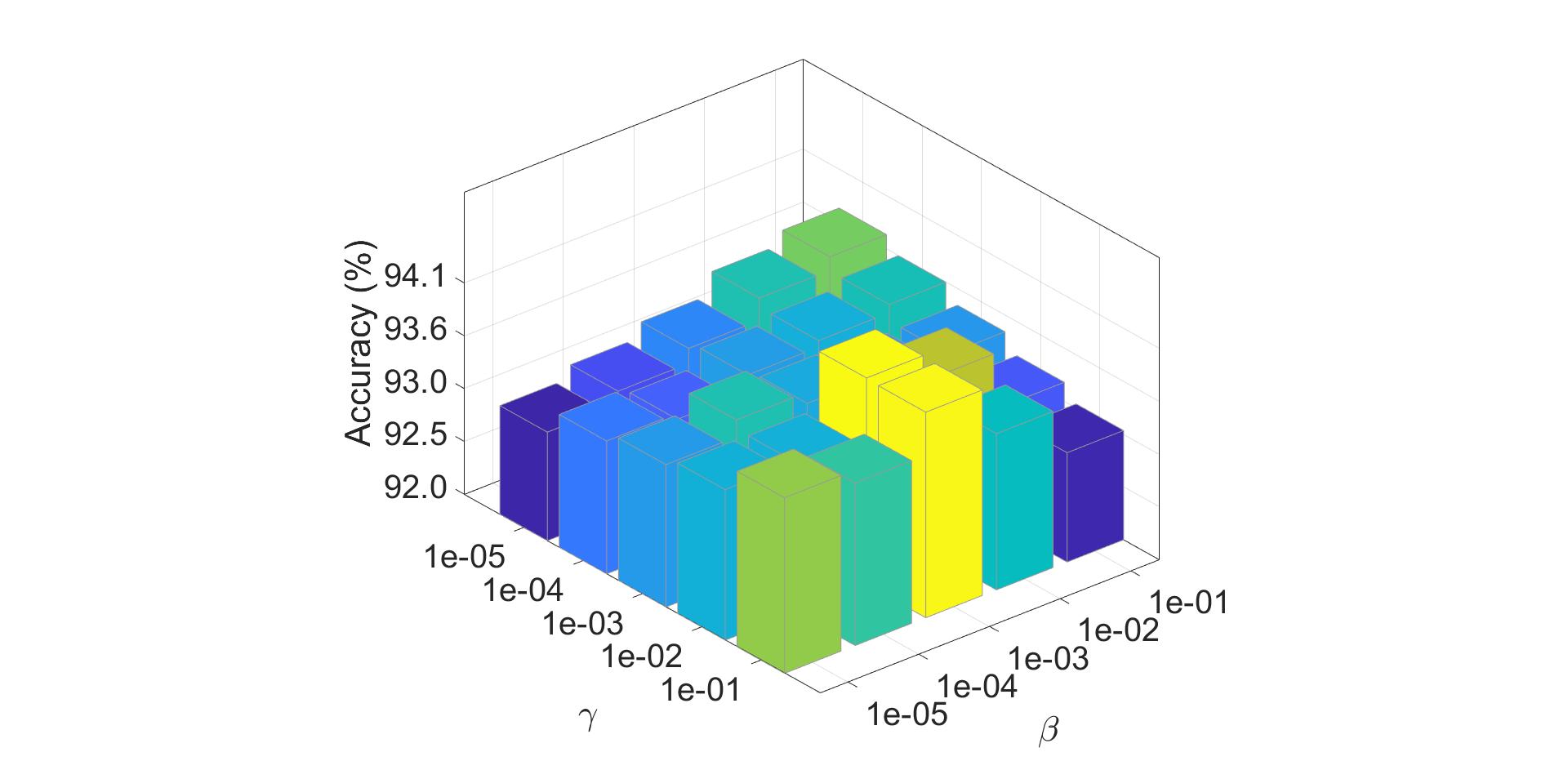}
    \caption{Stanford Dogs (ConvNeXt-tiny)}
    \label{fig:stanforddogs_convnext_gamma_comparison}
\end{figure}

\begin{figure}[!htbp]
    \centering
    \includegraphics[width=0.85\textwidth]{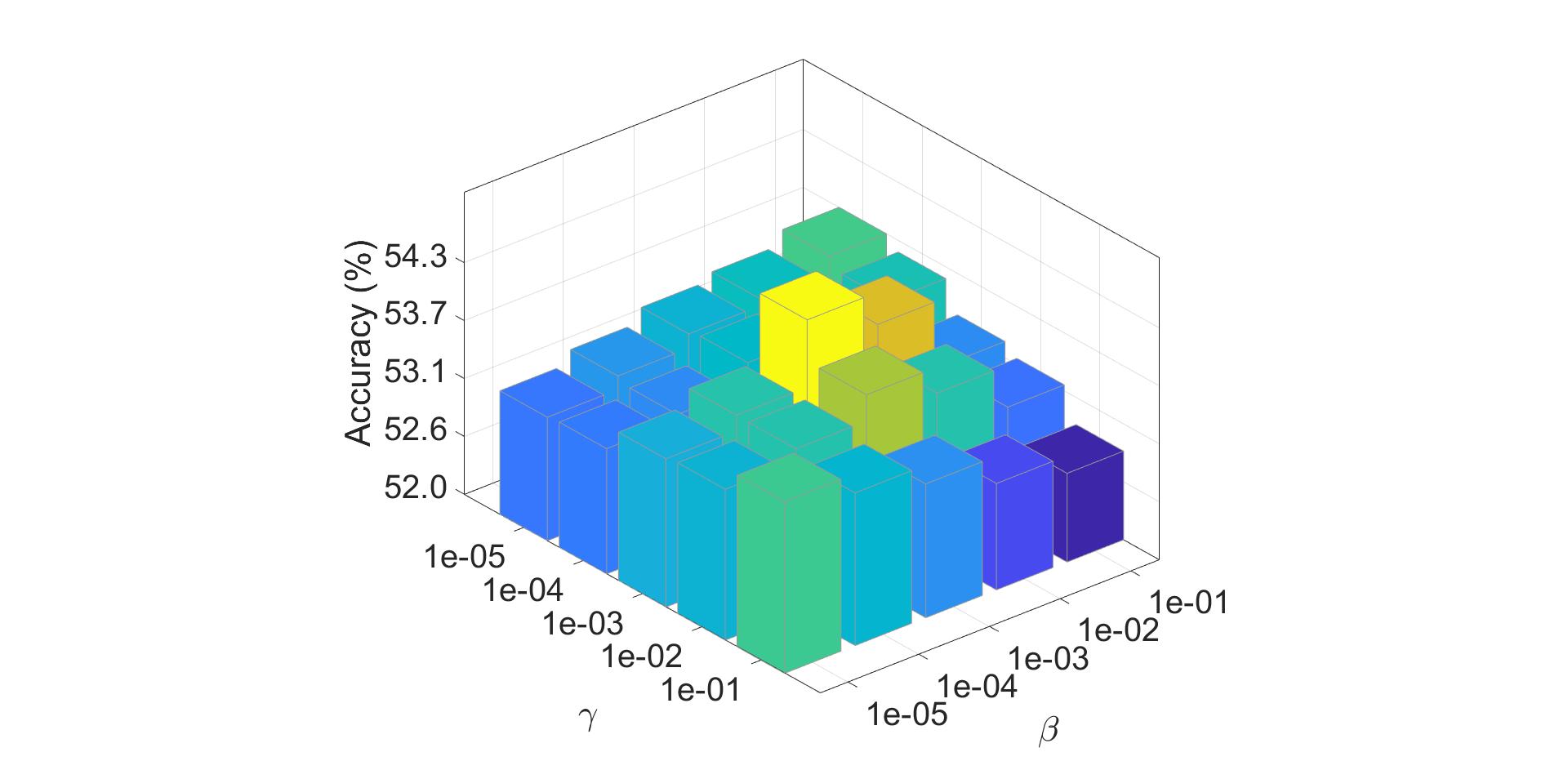}
    \caption{PASCAL VOC (TinyViT-5M)}
    \label{fig:pascalvoc_gamma_comparison}
\end{figure}

\clearpage 

\paragraph{Hyperparameters $\beta$ and $\gamma$:}
We further analyze the joint sensitivity of hyperparameters $\beta$ and $\gamma$. Figures~\ref{fig:cifar10_gamma_comparison} to~\ref{fig:pascalvoc_gamma_comparison} present heatmaps showing the accuracy under different combinations of $\beta$ and $\gamma$. We consistently observe optimal accuracy at $\beta = 1\times10^{-3}$ across all datasets. Interestingly, a clear difference emerges in optimal $\gamma$ values: general datasets (CIFAR-10, CIFAR-100, PASCAL VOC) achieve optimal results at $\gamma = 1\times10^{-3}$, whereas fine-grained datasets (CUB200-2011, Stanford Dogs) require a higher $\gamma = 1\times10^{-2}$, indicating that stricter orthogonality is essential for subtle intra-class feature distinctions.

These findings clearly illustrate the robustness of our parameter selection and provide practical guidelines for effective implementation of OSCS-SupCon across diverse datasets.

\subsection{Ablation Study}

Table~\ref{tab:ablation} summarizes the ablation experiments conducted to evaluate the contributions of individual components of our proposed OSCS-SupCon method across different datasets, using ResNet-18 as the backbone encoder. Specifically, we investigate four variants: (1) \textit{Sigmoid-only} ($\beta=0, \gamma=0$, omitting the style distance and orthogonality losses), (2) \textit{Sigmoid + $\beta$} ($\gamma=0$,  omitting only the orthogonality loss), (3) \textit{Sigmoid + $\gamma$} ($\beta=0$,   (omitting only the style distance loss) , and (4) the complete OSCS-SupCon variant (\textit{Sigmoid + $\beta$ + $\gamma$}).

We observe that our complete OSCS-SupCon consistently achieves the highest accuracy across all datasets. Notably, the \textit{Sigmoid-only} variant already demonstrates strong performance compared to classical SupCon, SelfCon, and CS-SupCon methods, clearly indicating the effectiveness of explicitly modeling pairwise relationships via our sigmoid-based loss, which significantly mitigates negative-sample dilution inherent in the InfoNCE-based losses.

Introducing either the explicit style-distance constraint ($\beta$ term) or the orthogonality constraint ($\gamma$ term) independently yields clear performance improvements, confirming their effectiveness in promoting intra-class style variability and enforcing subspace orthogonality, respectively. Furthermore, the integration of both constraints (Sigmoid + $\beta$ + $\gamma$) provides additional and consistent accuracy gains on all datasets, validating their complementary roles in jointly enhancing feature discriminability and robustness.

In summary, the ablation results strongly highlight the complementary effects of all three components in our OSCS-SupCon loss function, leading to significantly improved feature disentanglement, robustness, and generalization performance.

\begin{table*}[htbp]
\caption{Ablation study of OSCS-SupCon components with ResNet-18 backbone.}
\label{tab:ablation}
\centering
\resizebox{\textwidth}{!}{
\begin{tabular}{lcccccccc}
\toprule
Dataset & SupCon & SelfCon & CS-SupCon & CS-SupCon w. ov. & Sigmoid-only & Sigmoid+$\beta$ & Sigmoid+$\gamma$ & Sigmoid+$\beta$+$\gamma$ (Full) \\
 & & & & & ($\beta=0, \gamma=0$) & ($\gamma=0$) & ($\beta=0$) & (Ours) \\
\midrule
CIFAR10 & 94.7 & 95.1 & 94.9 & 95.2 & 95.2 & 95.4 & 95.5 & \textbf{95.6} \\
CIFAR100 & 73.6 & 74.9 & 74.1 & 75.2 & 75.3 & 75.6 & 75.6 & \textbf{75.8} \\
Tiny-ImageNet & 57.7 & 59.8 & 59.0 & 59.9 & 59.8 & 60.5 & 60.7 & \textbf{61.1} \\
CUB200-2011 & 57.1 & 60.4 & 59.1 & 61.0 & 60.8 & 61.7 & 62.1 & \textbf{62.5} \\
Stanford Dogs & 63.0 & 65.4 & 64.8 & 66.3 & 66.4 & 67.1 & 67.6 & \textbf{68.2}  \\
\bottomrule
\end{tabular}}
\end{table*}

\section{Conclusion}
\label{sec6}

In this paper, we introduced a novel feature disentanglement framework named Orthogonal Sigmoid-based Common and Style Supervised Contrastive Learning (OSCS-SupCon). Our proposed approach explicitly integrates a sigmoid-based pairwise contrastive loss into supervised contrastive learning and simultaneously enforces strict orthogonality constraints between common (category-relevant) and style (category-irrelevant) feature subspaces. Extensive experiments conducted on six benchmark image datasets, using multiple backbone encoders including CNNs and Transformers, clearly demonstrate that OSCS-SupCon consistently outperforms state-of-the-art methods, such as classical SupCon, CS-SupCon, and SelfCon. Ablation studies and rigorous statistical analyses further validate the effectiveness, reliability, and robustness of each component within our proposed framework.

The primary advantages of OSCS-SupCon are summarized as follows:  
(1) \textbf{Enhanced Negative-Sample Differentiation}: By explicitly mitigating negative-sample dilution through the sigmoid-based contrastive loss with adaptive learnable parameters, OSCS-SupCon significantly improves discriminative feature representation, especially beneficial for fine-grained classification tasks.  
(2) \textbf{Robust Style Variation Control}: The explicit style-distance constraint effectively manages intra-class style variations, further improving model robustness and generalization performance.  
(3) \textbf{Explicit Orthogonality Constraint}: The strict orthogonality constraints explicitly imposed between the common and style subspaces significantly reduce residual coupling between features, enhancing interpretability and the quality of feature disentanglement.

Despite these compelling advantages, OSCS-SupCon also introduces certain limitations, notably a degree of sensitivity to hyperparameter tuning (\(\beta\) and \(\gamma\)), particularly for fine-grained classification scenarios, which require careful parameter selection to achieve optimal performance.

In future work, we aim to further enhance OSCS-SupCon by developing adaptive strategies for automatic hyperparameter tuning, thereby improving its practical ease of use and stability across diverse tasks. Additionally, we plan to investigate extensions of our framework to semi-supervised, incremental, and transfer learning scenarios, broadening the applicability and practical value of our proposed method in diverse real-world applications.


\begin{thebibliography}{31}
\ifx \bisbn   \undefined \def \bisbn  #1{ISBN #1}\fi
\ifx \binits  \undefined \def \binits#1{#1}\fi
\ifx \bauthor  \undefined \def \bauthor#1{#1}\fi
\ifx \batitle  \undefined \def \batitle#1{#1}\fi
\ifx \bjtitle  \undefined \def \bjtitle#1{#1}\fi
\ifx \bvolume  \undefined \def \bvolume#1{\textbf{#1}}\fi
\ifx \byear  \undefined \def \byear#1{#1}\fi
\ifx \bissue  \undefined \def \bissue#1{#1}\fi
\ifx \bfpage  \undefined \def \bfpage#1{#1}\fi
\ifx \blpage  \undefined \def \blpage #1{#1}\fi
\ifx \burl  \undefined \def \burl#1{\textsf{#1}}\fi
\ifx \doiurl  \undefined \def \doiurl#1{\url{https://doi.org/#1}}\fi
\ifx \betal  \undefined \def \betal{\textit{et al.}}\fi
\ifx \binstitute  \undefined \def \binstitute#1{#1}\fi
\ifx \binstitutionaled  \undefined \def \binstitutionaled#1{#1}\fi
\ifx \bctitle  \undefined \def \bctitle#1{#1}\fi
\ifx \beditor  \undefined \def \beditor#1{#1}\fi
\ifx \bpublisher  \undefined \def \bpublisher#1{#1}\fi
\ifx \bbtitle  \undefined \def \bbtitle#1{#1}\fi
\ifx \bedition  \undefined \def \bedition#1{#1}\fi
\ifx \bseriesno  \undefined \def \bseriesno#1{#1}\fi
\ifx \blocation  \undefined \def \blocation#1{#1}\fi
\ifx \bsertitle  \undefined \def \bsertitle#1{#1}\fi
\ifx \bsnm \undefined \def \bsnm#1{#1}\fi
\ifx \bsuffix \undefined \def \bsuffix#1{#1}\fi
\ifx \bparticle \undefined \def \bparticle#1{#1}\fi
\ifx \barticle \undefined \def \barticle#1{#1}\fi
\ifx \bconfdate \undefined \def \bconfdate #1{#1}\fi
\ifx \botherref \undefined \def \botherref #1{#1}\fi
\ifx \bchapter \undefined \def \bchapter#1{#1}\fi
\ifx \bbook \undefined \def \bbook#1{#1}\fi
\ifx \bcomment \undefined \def \bcomment#1{#1}\fi
\ifx \oauthor \undefined \def \oauthor#1{#1}\fi
\ifx \citeauthoryear \undefined \def \citeauthoryear#1{#1}\fi
\ifx \endbibitem  \undefined \def \endbibitem {}\fi
\ifx \bconflocation  \undefined \def \bconflocation#1{#1}\fi
\ifx \arxivurl  \undefined \def \arxivurl#1{\textsf{#1}}\fi
\csname PreBibitemsHook\endcsname

\bibitem{DARBAN2025}
\begin{barticle}
\bauthor{\bsnm{Darban}, \binits{Z.Z.}},
\bauthor{\bsnm{Webb}, \binits{G.I.}},
\bauthor{\bsnm{Pan}, \binits{S.}},
\bauthor{\bsnm{Aggarwal}, \binits{C.C.}},
\bauthor{\bsnm{Salehi}, \binits{M.}}:
\batitle{Carla: Self-supervised contrastive representation learning for time series anomaly detection}.
\bjtitle{Pattern Recognition}
\bvolume{157},
\bfpage{110874}
(\byear{2025})
\end{barticle}
\endbibitem

\bibitem{pmlr-v162-deng22c}
\begin{bchapter}
\bauthor{\bsnm{Deng}, \binits{X.}},
\bauthor{\bsnm{Zhang}, \binits{Z.}}:
\bctitle{Deep causal metric learning}.
In: \bbtitle{Proceedings of the 39th International Conference on Machine Learning}.
\bsertitle{Proceedings of Machine Learning Research},
vol. \bseriesno{162},
pp. \bfpage{4993}--\blpage{5006}
(\byear{2022})
\end{bchapter}
\endbibitem

\bibitem{Bao2023}
\begin{barticle}
\bauthor{\bsnm{Bao}, \binits{S.}},
\bauthor{\bsnm{Xu}, \binits{Q.}},
\bauthor{\bsnm{Zhiyong~Yang}, \binits{X.C.}},
\bauthor{\bsnm{Huang}, \binits{Q.}}:
\batitle{Rethinking collaborative metric learning: Toward an efficient alternative without negative sampling}.
\bjtitle{IEEE Transactions on Pattern Analysis and Machine Intelligence}
\bvolume{45},
\bfpage{1017}--\blpage{1035}
(\byear{2023})
\end{barticle}
\endbibitem

\bibitem{pmlr-v139-roth21a}
\begin{bchapter}
\bauthor{\bsnm{Roth}, \binits{K.}},
\bauthor{\bsnm{Milbich}, \binits{T.}},
\bauthor{\bsnm{Ommer}, \binits{B.}},
\bauthor{\bsnm{Cohen}, \binits{J.P.}},
\bauthor{\bsnm{Ghassemi}, \binits{M.}}:
\bctitle{Simultaneous similarity-based self-distillation for deep metric learning}.
In: \beditor{\bsnm{Meila}, \binits{M.}},
\beditor{\bsnm{Zhang}, \binits{T.}} (eds.)
\bbtitle{Proceedings of the 38th International Conference on Machine Learning}.
\bsertitle{Proceedings of Machine Learning Research},
vol. \bseriesno{139},
pp. \bfpage{9095}--\blpage{9106}
(\byear{2021})
\end{bchapter}
\endbibitem

\bibitem{zhu2021visual}
\begin{botherref}
\oauthor{\bsnm{Zhu}, \binits{S.}},
\oauthor{\bsnm{Yang}, \binits{T.}},
\oauthor{\bsnm{Chen}, \binits{C.}}:
Visual explanation for deep metric learning.
IEEE Transactions on Image Processing
(2021)
\end{botherref}
\endbibitem

\bibitem{Gonzalez-Zapata_2022_CVPR}
\begin{bchapter}
\bauthor{\bsnm{Gonzalez-Zapata}, \binits{J.}},
\bauthor{\bsnm{Reyes-Amezcua}, \binits{I.}},
\bauthor{\bsnm{Flores-Araiza}, \binits{D.}},
\bauthor{\bsnm{Mendez-Ruiz}, \binits{M.}},
\bauthor{\bsnm{Ochoa-Ruiz}, \binits{G.}},
\bauthor{\bsnm{Mendez-Vazquez}, \binits{A.}}:
\bctitle{Guided deep metric learning}.
In: \bbtitle{Proceedings of the IEEE/CVF Conference on Computer Vision and Pattern Recognition (CVPR) Workshops},
pp. \bfpage{1481}--\blpage{1489}
(\byear{2022})
\end{bchapter}
\endbibitem

\bibitem{WANG2024}
\begin{barticle}
\bauthor{\bsnm{Wang}, \binits{J.}},
\bauthor{\bsnm{Pang}, \binits{Y.}},
\bauthor{\bsnm{Cao}, \binits{J.}},
\bauthor{\bsnm{Sun}, \binits{H.}},
\bauthor{\bsnm{Shao}, \binits{Z.}},
\bauthor{\bsnm{Li}, \binits{X.}}:
\batitle{Deep intra-image contrastive learning for weakly supervised one-step person search}.
\bjtitle{Pattern Recognition}
\bvolume{147},
\bfpage{110047}
(\byear{2024}).
\doiurl{10.1016/j.patcog.2023.110047}
\end{barticle}
\endbibitem

\bibitem{Wang2022}
\begin{barticle}
\bauthor{\bsnm{Wang}, \binits{Y.}},
\bauthor{\bsnm{Liu}, \binits{P.}},
\bauthor{\bsnm{Lang}, \binits{Y.}},
\bauthor{\bsnm{Zhou}, \binits{Q.}},
\bauthor{\bsnm{Shan}, \binits{X.}}:
\batitle{Learnable dynamic margin in deep metric learning}.
\bjtitle{Pattern Recognition}
\bvolume{132},
\bfpage{108961}
(\byear{2022}).
\doiurl{10.1016/j.patcog.2022.108961}
\end{barticle}
\endbibitem

\bibitem{khosla2020supervised}
\begin{barticle}
\bauthor{\bsnm{Khosla}, \binits{P.}},
\bauthor{\bsnm{Teterwak}, \binits{P.}},
\bauthor{\bsnm{Wang}, \binits{C.}},
\bauthor{\bsnm{Sarna}, \binits{A.}},
\bauthor{\bsnm{Tian}, \binits{Y.}},
\bauthor{\bsnm{Isola}, \binits{P.}},
\bauthor{\bsnm{Maschinot}, \binits{A.}},
\bauthor{\bsnm{Liu}, \binits{C.}},
\bauthor{\bsnm{Krishnan}, \binits{D.}}:
\batitle{Supervised contrastive learning}.
\bjtitle{Advances in neural information processing systems}
\bvolume{33},
\bfpage{18661}--\blpage{18673}
(\byear{2020})
\end{barticle}
\endbibitem

\bibitem{Li2023}
\begin{barticle}
\bauthor{\bsnm{Li}, \binits{X.}},
\bauthor{\bsnm{Yang}, \binits{X.}},
\bauthor{\bsnm{Ma}, \binits{Z.}},
\bauthor{\bsnm{Xue}, \binits{J.-H.}}:
\batitle{Deep metric learning for few-shot image classification: A review of recent developments}.
\bjtitle{Pattern Recognition}
\bvolume{138},
\bfpage{109381}
(\byear{2023}).
\doiurl{10.1016/j.patcog.2023.109381}
\end{barticle}
\endbibitem

\bibitem{ZHANG2023}
\begin{barticle}
\bauthor{\bsnm{Zhang}, \binits{J.}},
\bauthor{\bsnm{Zhang}, \binits{X.-Y.}},
\bauthor{\bsnm{Wang}, \binits{C.}},
\bauthor{\bsnm{Liu}, \binits{C.-L.}}:
\batitle{Deep representation learning for domain generalization with information bottleneck principle}.
\bjtitle{Pattern Recognition}
\bvolume{143},
\bfpage{109737}
(\byear{2023}).
\doiurl{10.1016/j.patcog.2023.109737}
\end{barticle}
\endbibitem

\bibitem{oord2018representation}
\begin{botherref}
\oauthor{\bsnm{Oord}, \binits{A.v.d.}},
\oauthor{\bsnm{Li}, \binits{Y.}},
\oauthor{\bsnm{Vinyals}, \binits{O.}}:
Representation learning with contrastive predictive coding.
arXiv preprint arXiv:1807.03748
(2018)
\end{botherref}
\endbibitem

\bibitem{WU2018}
\begin{barticle}
\bauthor{\bsnm{Wu}, \binits{B.}},
\bauthor{\bsnm{Chen}, \binits{Z.}},
\bauthor{\bsnm{Wang}, \binits{J.}},
\bauthor{\bsnm{Wu}, \binits{H.}}:
\batitle{Exponential discriminative metric embedding in deep learning}.
\bjtitle{Neurocomputing}
\bvolume{290},
\bfpage{108}--\blpage{120}
(\byear{2018})
\end{barticle}
\endbibitem

\bibitem{Sohn2016}
\begin{botherref}
\oauthor{\bsnm{Sohn}, \binits{K.}}:
Improved deep metric learning with multi-class n-pair loss objective.
In: Advances in Neural Information Processing Systems 29,
Red Hook, NY, USA,
pp. 1857--1865
\end{botherref}
\endbibitem

\bibitem{Wang2019}
\begin{bchapter}
\bauthor{\bsnm{Wang}, \binits{X.}},
\bauthor{\bsnm{Han}, \binits{X.}},
\bauthor{\bsnm{Huang}, \binits{W.}},
\bauthor{\bsnm{Dong}, \binits{D.}},
\bauthor{\bsnm{Scott}, \binits{M.R.}}:
\bctitle{Multi-similarity loss with general pair weighting for deep metric learning}.
In: \bbtitle{Proceedings of the IEEE Conference on Computer Vision and Pattern Recognition},
pp. \bfpage{5022}--\blpage{5030}
(\byear{2019})
\end{bchapter}
\endbibitem

\bibitem{Chen2020}
\begin{bchapter}
\bauthor{\bsnm{Chen}, \binits{T.}},
\bauthor{\bsnm{Kornblith}, \binits{S.}},
\bauthor{\bsnm{Norouzi}, \binits{M.}},
\bauthor{\bsnm{Hinton}, \binits{G.}}:
\bctitle{A simple framework for contrastive learning of visual representations}.
In: \bbtitle{Proceedings of the 37th International Conference on Machine Learning}.
\bsertitle{ICML'20}
(\byear{2020})
\end{bchapter}
\endbibitem

\bibitem{Grill2020}
\begin{bchapter}
\bauthor{\bsnm{Grill}, \binits{J.-B.}},
\bauthor{\bsnm{Strub}, \binits{F.}},
\bauthor{\bsnm{Altch\'{e}}, \binits{F.}},
\bauthor{\bsnm{Tallec}, \binits{C.}},
\bauthor{\bsnm{Richemond}, \binits{P.H.}},
\bauthor{\bsnm{Buchatskaya}, \binits{E.}},
\bauthor{\bsnm{Doersch}, \binits{C.}},
\bauthor{\bsnm{Pires}, \binits{B.A.}},
\bauthor{\bsnm{Guo}, \binits{Z.D.}},
\bauthor{\bsnm{Azar}, \binits{M.G.}},
\bauthor{\bsnm{Piot}, \binits{B.}},
\bauthor{\bsnm{Kavukcuoglu}, \binits{K.}},
\bauthor{\bsnm{Munos}, \binits{R.}},
\bauthor{\bsnm{Valko}, \binits{M.}}:
\bctitle{Bootstrap your own latent a new approach to self-supervised learning}.
In: \bbtitle{Proceedings of the 34th International Conference on Neural Information Processing Systems}.
\bsertitle{NIPS'20}
(\byear{2020})
\end{bchapter}
\endbibitem

\bibitem{dornaika2025deep}
\begin{barticle}
\bauthor{\bsnm{Dornaika}, \binits{F.}},
\bauthor{\bsnm{Wang}, \binits{B.}},
\bauthor{\bsnm{Charafeddine}, \binits{J.}}:
\batitle{Deep feature disentanglement for supervised contrastive learning: Application to image classification}.
\bjtitle{Cognitive Computation}
\bvolume{17}(\bissue{3}),
\bfpage{1}--\blpage{12}
(\byear{2025})
\end{barticle}
\endbibitem

\bibitem{baeSelfContrastiveLearningSingleviewed2022}
\begin{bchapter}
\bauthor{\bsnm{Bae}, \binits{S.}},
\bauthor{\bsnm{Kim}, \binits{S.}},
\bauthor{\bsnm{Ko}, \binits{J.}},
\bauthor{\bsnm{Lee}, \binits{G.}},
\bauthor{\bsnm{Noh}, \binits{S.}},
\bauthor{\bsnm{Yun}, \binits{S.-Y.}}:
\bctitle{Self-contrastive learning: single-viewed supervised contrastive framework using sub-network}.
In: \bbtitle{Proceedings of the AAAI Conference on Artificial Intelligence},
vol. \bseriesno{37},
pp. \bfpage{197}--\blpage{205}
(\byear{2023})
\end{bchapter}
\endbibitem

\bibitem{LIU2023}
\begin{barticle}
\bauthor{\bsnm{Liu}, \binits{Z.}},
\bauthor{\bsnm{Wu}, \binits{F.}},
\bauthor{\bsnm{Wang}, \binits{Y.}},
\bauthor{\bsnm{Yang}, \binits{M.}},
\bauthor{\bsnm{Pan}, \binits{X.}}:
\batitle{Fedcl: Federated contrastive learning for multi-center medical image classification}.
\bjtitle{Pattern Recognition}
\bvolume{143},
\bfpage{109739}
(\byear{2023}).
\doiurl{10.1016/j.patcog.2023.109739}
\end{barticle}
\endbibitem

\bibitem{HU2024}
\begin{barticle}
\bauthor{\bsnm{Hu}, \binits{H.}},
\bauthor{\bsnm{Wang}, \binits{X.}},
\bauthor{\bsnm{Zhang}, \binits{Y.}},
\bauthor{\bsnm{Chen}, \binits{Q.}},
\bauthor{\bsnm{Guan}, \binits{Q.}}:
\batitle{A comprehensive survey on contrastive learning}.
\bjtitle{Neurocomputing}
\bvolume{610},
\bfpage{128645}
(\byear{2024})
\end{barticle}
\endbibitem

\bibitem{Cheng2021}
\begin{bchapter}
\bauthor{\bsnm{Tan}, \binits{C.}},
\bauthor{\bsnm{Xia}, \binits{J.}},
\bauthor{\bsnm{Wu}, \binits{L.}},
\bauthor{\bsnm{Li}, \binits{S.Z.}}:
\bctitle{Co-learning: Learning from noisy labels with self-supervision}.
In: \bbtitle{Proceedings of the 29th ACM International Conference on Multimedia}.
\bsertitle{MM '21},
pp. \bfpage{1405}--\blpage{1413}
(\byear{2021})
\end{bchapter}
\endbibitem

\bibitem{Chen2021}
\begin{bchapter}
\bauthor{\bsnm{Chen}, \binits{D.}},
\bauthor{\bsnm{Chen}, \binits{Y.}},
\bauthor{\bsnm{Li}, \binits{Y.}},
\bauthor{\bsnm{Mao}, \binits{F.}},
\bauthor{\bsnm{He}, \binits{Y.}},
\bauthor{\bsnm{Xue}, \binits{H.}}:
\bctitle{Self-supervised learning for few-shot image classification}.
In: \bbtitle{IEEE International Conference on Acoustics, Speech and Signal Processing (ICASSP)},
pp. \bfpage{1745}--\blpage{1749}
(\byear{2021})
\end{bchapter}
\endbibitem

\bibitem{Henaff2020}
\begin{bchapter}
\bauthor{\bsnm{Hénaff}, \binits{O.J.}},
\bauthor{\bsnm{Srinivas}, \binits{A.}},
\bauthor{\bsnm{De~Fauw}, \binits{J.}},
\bauthor{\bsnm{Razavi}, \binits{A.}},
\bauthor{\bsnm{Doersch}, \binits{C.}},
\bauthor{\bsnm{Eslami}, \binits{S.M.A.}},
\bauthor{\bsnm{Oord}, \binits{A.v.d.}}:
\bctitle{Data-efficient image recognition with contrastive predictive coding}.
In: \bbtitle{International Conference on Machine Learning}
(\byear{2020})
\end{bchapter}
\endbibitem

\bibitem{ZHAO2024}
\begin{barticle}
\bauthor{\bsnm{Zhao}, \binits{T.}},
\bauthor{\bsnm{Lin}, \binits{Y.}},
\bauthor{\bsnm{Wu}, \binits{Y.}},
\bauthor{\bsnm{Du}, \binits{B.}}:
\batitle{Promote knowledge mining towards open-world semi-supervised learning}.
\bjtitle{Pattern Recognition}
\bvolume{149},
\bfpage{110259}
(\byear{2024}).
\doiurl{10.1016/j.patcog.2024.110259}
\end{barticle}
\endbibitem

\bibitem{xue2023features}
\begin{bchapter}
\bauthor{\bsnm{Xue}, \binits{Y.}},
\bauthor{\bsnm{Joshi}, \binits{S.}},
\bauthor{\bsnm{Gan}, \binits{E.}},
\bauthor{\bsnm{Chen}, \binits{P.-Y.}},
\bauthor{\bsnm{Mirzasoleiman}, \binits{B.}}:
\bctitle{Which features are learnt by contrastive learning? on the role of simplicity bias in class collapse and feature suppression}.
In: \bbtitle{International Conference on Machine Learning},
pp. \bfpage{38938}--\blpage{38970}
(\byear{2023}).
\bcomment{PMLR}
\end{bchapter}
\endbibitem

\bibitem{azizi2023robust}
\begin{barticle}
\bauthor{\bsnm{Azizi}, \binits{S.}},
\bauthor{\bsnm{Culp}, \binits{L.}},
\bauthor{\bsnm{Freyberg}, \binits{J.}},
\bauthor{\bsnm{Mustafa}, \binits{B.}},
\bauthor{\bsnm{Baur}, \binits{S.}},
\bauthor{\bsnm{Kornblith}, \binits{S.}},
\bauthor{\bsnm{Chen}, \binits{T.}},
\bauthor{\bsnm{Tomasev}, \binits{N.}},
\bauthor{\bsnm{Mitrovi{\'c}}, \binits{J.}},
\bauthor{\bsnm{Strachan}, \binits{P.}}, \betal:
\batitle{Robust and data-efficient generalization of self-supervised machine learning for diagnostic imaging}.
\bjtitle{Nature Biomedical Engineering}
\bvolume{7}(\bissue{6}),
\bfpage{756}--\blpage{779}
(\byear{2023})
\end{barticle}
\endbibitem

\bibitem{zhang2024hierarchical}
\begin{barticle}
\bauthor{\bsnm{Zhang}, \binits{J.}},
\bauthor{\bsnm{Li}, \binits{Y.}},
\bauthor{\bsnm{Shen}, \binits{F.}},
\bauthor{\bsnm{He}, \binits{Y.}},
\bauthor{\bsnm{Tan}, \binits{H.}},
\bauthor{\bsnm{He}, \binits{Y.}}:
\batitle{Hierarchical text classification with multi-label contrastive learning and knn}.
\bjtitle{Neurocomputing}
\bvolume{577},
\bfpage{127323}
(\byear{2024})
\end{barticle}
\endbibitem

\bibitem{DUAN2025}
\begin{barticle}
\bauthor{\bsnm{Duan}, \binits{C.}},
\bauthor{\bsnm{Jiao}, \binits{Y.}},
\bauthor{\bsnm{Kang}, \binits{L.}},
\bauthor{\bsnm{Yang}, \binits{J.Z.}},
\bauthor{\bsnm{Zhou}, \binits{F.}}:
\batitle{Deep contrastive representation learning for supervised tasks}.
\bjtitle{Pattern Recognition}
\bvolume{161},
\bfpage{111309}
(\byear{2025})
\end{barticle}
\endbibitem

\bibitem{He15_resnet50}
\begin{botherref}
\oauthor{\bsnm{He}, \binits{K.}},
\oauthor{\bsnm{Zhang}, \binits{X.}},
\oauthor{\bsnm{Ren}, \binits{S.}},
\oauthor{\bsnm{Sun}, \binits{J.}}:
Deep residual learning for image recognition.
CoRR
\textbf{abs/1512.03385}
(2015)
{\href{https://arxiv.org/abs/1512.03385}{{arXiv:1512.03385}}}
\end{botherref}
\endbibitem

\bibitem{liu2022convnet}
\begin{botherref}
\oauthor{\bsnm{Liu}, \binits{Z.}},
\oauthor{\bsnm{Mao}, \binits{H.}},
\oauthor{\bsnm{Wu}, \binits{C.-Y.}},
\oauthor{\bsnm{Feichtenhofer}, \binits{C.}},
\oauthor{\bsnm{Darrell}, \binits{T.}},
\oauthor{\bsnm{Xie}, \binits{S.}}:
A convnet for the 2020s.
Proceedings of the IEEE/CVF Conference on Computer Vision and Pattern Recognition (CVPR)
(2022)
\end{botherref}
\endbibitem

\end{thebibliography}

\end{document}